\pdfoutput=1

\documentclass[11pt]{article}

\usepackage[preprint]{acl}
\usepackage{amssymb}

\usepackage{times}
\usepackage{latexsym}
\usepackage{enumitem}
\usepackage{listings}
\usepackage{algorithm}
\usepackage{algorithmic}
\usepackage{caption, subcaption}
\usepackage{booktabs} 

\usepackage{amsmath}
\usepackage{multirow}
\usepackage{longtable}

\usepackage[T1]{fontenc}

\usepackage[utf8]{inputenc}

\usepackage{microtype}

\usepackage{inconsolata}

\usepackage{graphicx}

\usepackage{soul}

\newcommand{\vx}{\mathbf{x}}

\newcommand{\vf}{\mathbf{f}}
\newcommand{\vW}{\mathbf{W}}
\newcommand{\vb}{\mathbf{b}}

\title{Feature-Level Insights into Artificial Text Detection with Sparse Autoencoders}
\author{
  \textbf{Kristian Kuznetsov\textsuperscript{1,2}},
  \textbf{Laida Kushnareva\textsuperscript{2}},
\textbf{Polina Druzhinina\textsuperscript{1,5}},
  \textbf{Anton Razzhigaev\textsuperscript{1,5}}, 
  \\
  \textbf{Anastasia Voznyuk\textsuperscript{3}},
  \textbf{Irina Piontkovskaya\textsuperscript{2}},
  \textbf{Evgeny Burnaev\textsuperscript{1,5}},
  \textbf{Serguei Barannikov\textsuperscript{1,4}},
\\
\\
  \textsuperscript{1}Skolkovo Institute of Science and Technology,
  \textsuperscript{2}AI Foundation and Algorithm Lab\\
  \textsuperscript{3}Moscow Institute of Physics and Technology,
  \textsuperscript{4}CNRS, Université Paris Cité, France\\
  \textsuperscript{5}Artificial Intelligence Research Institute (AIRI)
}

\begin{document}
\maketitle
\begin{abstract}

Artificial Text Detection (ATD) is becoming increasingly important with the rise of advanced Large Language Models (LLMs). Despite numerous efforts, no single algorithm performs consistently well across different types of unseen text or guarantees effective generalization to new LLMs. Interpretability plays a crucial role in achieving this goal. In this study, we enhance ATD interpretability by using Sparse Autoencoders (SAE) to extract features from Gemma-2-2b’s residual stream. We identify both interpretable and efficient features, analyzing their semantics and relevance through domain- and model-specific statistics, a steering approach, and manual or LLM-based interpretation. Our methods offer valuable insights into how texts from various models differ from human-written content. We show that modern LLMs have a distinct writing style, especially in information-dense domains, even though they can produce human-like outputs with personalized prompts.
\end{abstract}

\section{Introduction}

The active development of large language models (LLMs) has led to the increasing presence of AI-generated text in various domains, including news, education, and scientific literature. Although these models have demonstrated impressive fluency and coherence, concerns about misinformation, plagiarism, and AI-generated disinformation have required the development of reliable artificial text detection (ATD) systems  \cite{abdali2024decoding}. Existing ATD frameworks primarily rely on statistical measures, linguistic heuristics, and deep learning classifiers, yet these methods often lack interpretability, limiting their reliability in high-stakes applications \cite{yang-etal-2024-survey}.

\begin{figure}[t]
    \centering
    \includegraphics[width=\linewidth]{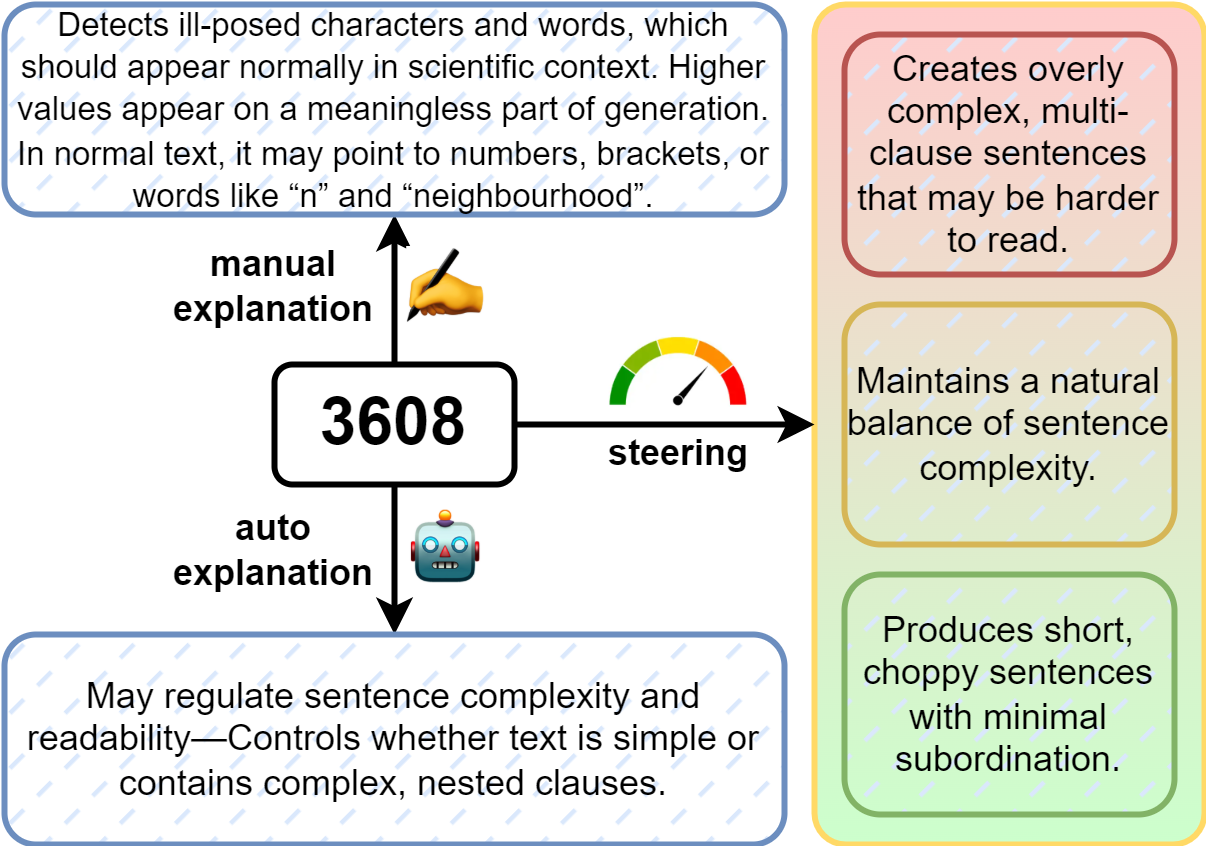}
    \caption{Interpretations of one of the most ``universal'' SAE features that are useful for ATD task.}
    \label{fig:example_feature_interpretation_3608}
\end{figure}

A promising approach to enhancing interpretability in ATD is the use of Sparse Autoencoders (SAEs), which learn structured representations of textual data by enforcing sparsity constraints \cite{huben2023sparse, makelov2024towards}. We can extract human-interpretable features that capture the underlying structure of text. 

In this study, we extend this line of research by applying SAEs from the Gemma-2-2b model \cite{gemmateam2024gemma2improvingopen} residual streams to analyze features that contribute to artificial text detection. By examining these features, we introduce a categorization of extracted features into discourse features (capturing long-range dependencies), noise features (highlighting unnatural artifacts), and style features (distinguishing stylistic variations). Our contributions are the following:

    (i) we demonstrate the efficiency of SAE for the ATD task; 
     (ii) we extract features which alone can effectively detect artificial texts for some domains and generation methods; 
    (iii) interpreting these features, we identify meaningful patterns that contribute to ATD interpretability.

For our main dataset, we utilized a highly comprehensive and up-to-date dataset from GenAI Content Detection Task 1—a shared task on binary machine-generated text detection, conducted as part of the GenAI workshop at COLING 2025 \cite{wang-etal-2025-genai}. Hereafter referred to as the COLING dataset, it contains a diverse range of model generations, from mT5 and OPT to GPT-4o and LLaMA-3. A complete list of models, along with generation examples, is provided in Appendix~\ref{sec:generation_examples}.

We also performed additional experiments on the RAID dataset \cite{dugan-etal-2024-raid}, which contains generations from several models with various sampling methods and a wide range of attacks, from paraphrasing to homoglyph-based modifications. We provide the full list of models and attacks, along with examples of generations, in Appendix~\ref{sec:generation_examples_raid}.

\section{Background}

Given a token sequence $(t_1, t_2, ..., t_n)$, an LLM computes hidden representations $\vx_i \in \mathbb{R}^d$ at each layer $l$ as
$\vx_i^{(l)} = g^{(l)} (\vx_1^{(l-1)}, \vx_2^{(l-1)},..., \vx_i^{(l-1)})$,
where $g$ represents a transformer block, typically including self-attention and feedforward operations. These activations encode meaningful information about text, but understanding models requires breaking them into analyzable features. Individual neurons are limited as features due to polysemanticity \cite{olah2020zoom}, meaning that models learn more semantic features than there are available dimensions in a layer; this situation is referred to as  superposition~\cite{elhage2022toy}.
To recover these features, a Sparse Autoencoder (SAE) has been proposed to identify a set of directions in activation space such that each activation vector is a sparse linear combination of them \cite{shakley2023superposition}.

Given activations $\vx$ from a language model, a sparse autoencoder decomposes and reconstructs them using encoder and decoder functions with some activation function $\sigma$:
$$f(\vx) = \sigma(\vW_{\text{enc}} \vx + \vb_{\text{enc}})$$
$$\hat{x}(f) = \vW_{\text{dec}} f(\vx) + \vb_{\text{dec}}$$
for which $\hat{x}(f(\vx))$ should map back to $\vx$. Here, the sparse and non-negative feature vector $f(\vx) \in \mathbb{R}^M$ (with $M \gg d$) specifies how to combine columns of $\vW_{\text{dec}}$ - learned features, or latents - to reconstruct $\vx$.

\section{Methods}
In this work, we take a step towards improving the interpretability of artificial text detection using SAEs. We employ the Gemma-2-2B model along with pre-trained autoencoders on residual streams from Gemma-Scope \cite{lieberum-etal-2024-gemma}.

\textbf{Classifier models.} For each even layer, we utilize an individual SAE  $(f^{(l)}, \hat{x}^{(l)})$ to extract learned features from each token. To obtain a feature vector $\vf$ representing the entire text for layer $l$, we sum over all tokens, yielding 
$$\vf = \sum_{i=1}^n f^{(l)}(\vx_i^{(l)})$$

We use an XGBoost classifier to evaluate the expressiveness of the full feature sets for each layer and identify the most important features for further analysis. The classifiers are trained exclusively on the Train subset of COLING and evaluated on the similar Dev set, as well as on the entirely distinct Devtest and Test subsets.

For a detailed feature analysis, we also use threshold classifiers on individual features.

\textbf{Manual Interpretation and Feature Steering.} 
For manual interpretation, we analyzed the texts that activate the most important features. In layers with strong performance and generalization (layers 8 to 20), we selected the top 20 most significant features identified by XGBoost, as well as all features that achieved the highest detection performance for each domain and model using a threshold classifier. The selected features, their statistical properties, and example texts are publicly available\footnote{\url{https://mgtsaevis.github.io/mgt-sae-visualization/}}.

To examine how learned features affect text generation, we use feature steering, which enables targeted modifications by selectively adjusting latent feature activations.
For a given feature with number $i$ associated with a specific text property, we first compute its maximum activation $A_{\max}$ across a reference dataset. During generation, hidden states are modified as
$$\vx' = \vx + \lambda A_{\max} \mathbf{d}_i$$
where $\vx$ is the original hidden state, $\mathbf{d}_i$ is the column of $\vW_{\text{dec}}$ and $\lambda$ is a scaling factor controlling the steering effect.
 
 Furthermore, we employed the GPT-4-o model to analyze changes across all sequences and determine the nature or function of a particular hidden feature. (see Appendix~\ref{sec:steering})

\section{Results}

\begin{figure*}[t] 
    \centering
    \includegraphics[width=\textwidth]{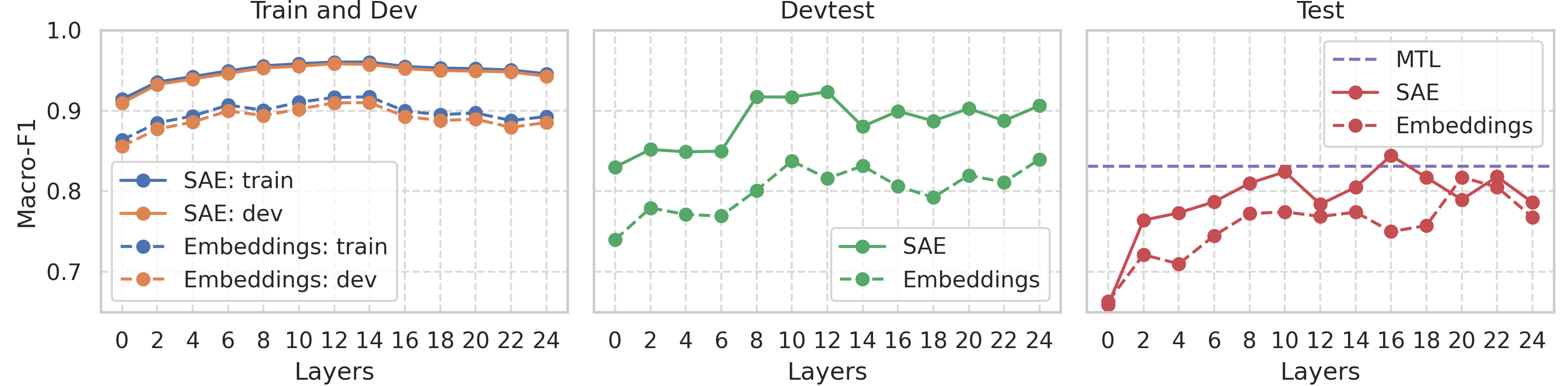} 
    \caption{Macro F1 for XGBoost model on activations and SAE-derived features on different subsets of COLING}
    \label{fig:main_metrics}
\end{figure*}

\begin{figure*}[t] 
    \centering
    \includegraphics[width=\textwidth]{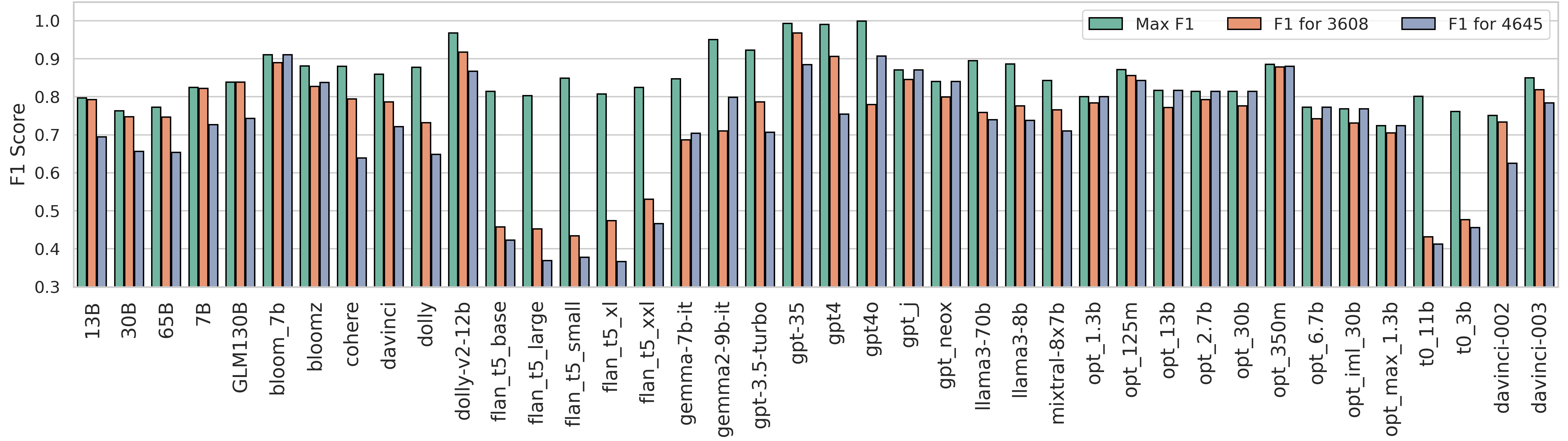} 
    \caption{Macro F1 for a threshold classifier on individual features across each model for the 16th layer. Max F1 presents the maximum F1 score for every feature;  features 3608 and 4645 are considered general features}
    \label{fig:models_bars}
\end{figure*}

\textbf{General Detection Quality.} To verify that SAE-derived features enable the detection of artificially generated texts, we apply XGBoost on these features and compare the results with XGBoost applied to mean-pooled activations from the layers. For training, we use the Train Subset, while testing is conducted on all remaining data.

As shown in Figure \ref{fig:main_metrics}, both SAE features and activations perform well on this subset but degrade slightly on others. Notably, SAE features outperform activations both in training and across other subsets, suggesting that removing superposition helps the classifier focus on more fundamental, atomic features.

Although our primary objective is interpretability, it is worth noting that, at the 16th layer, SAE-derived features outperform the state-of-the-art MTL model on this dataset \cite{gritsai-etal-2025-advacheck}.

\begin{figure}[t] 
    \centering
    \includegraphics[width=\linewidth]{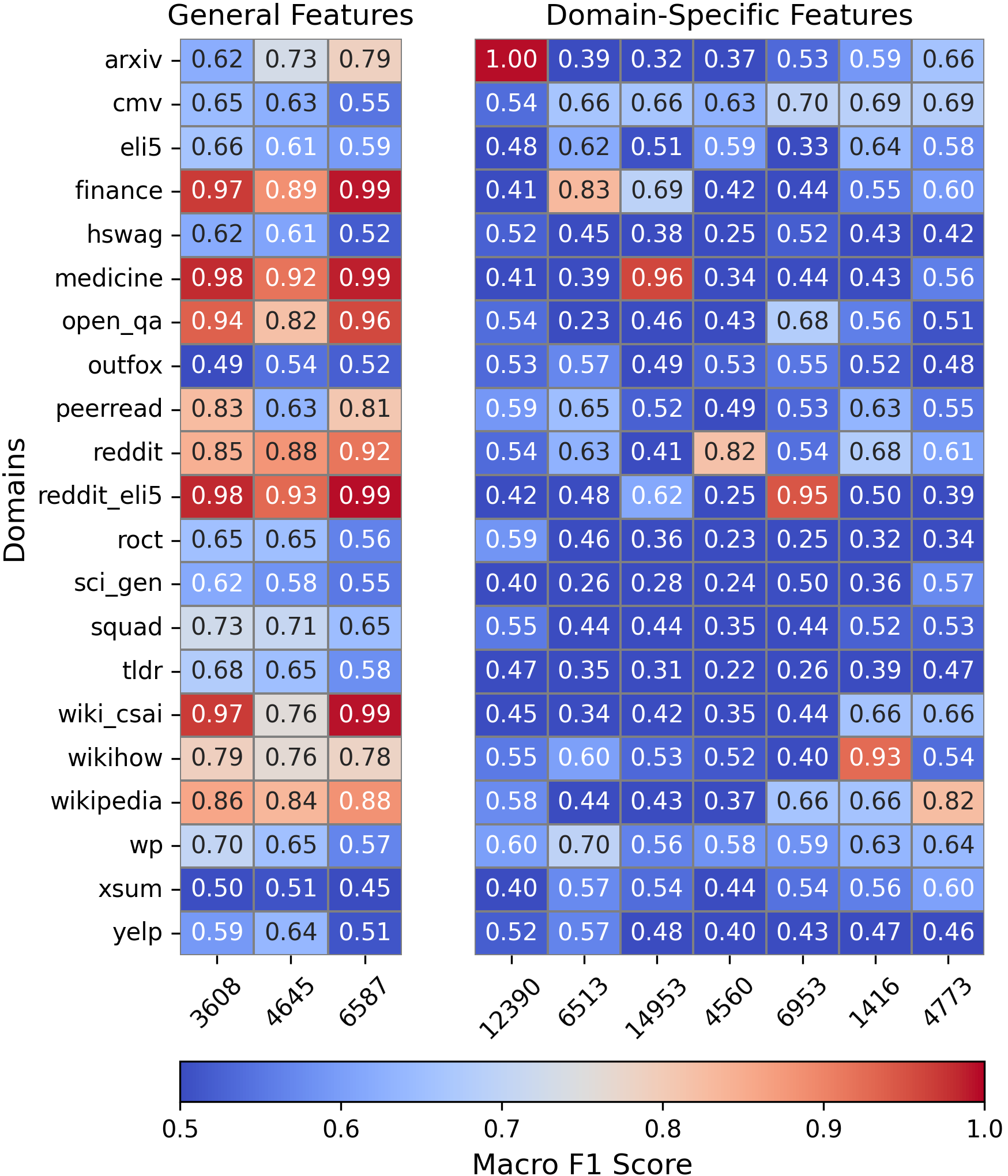} 
    \caption{F1 Macro by the domains subsets for some general and domain-specific features for the 16 layer}
    \label{fig:domains_heatmap}
\end{figure}

\noindent\textbf{Domain/Model-Specific and General Features.} In our analysis of feature structure, we aim to distinguish between general features and domain- or model-specific features. Our focus is on the 16th layer, as its features have proven to be the most expressive and lead to the best generalization, as discussed in the previous section. Given the highly imbalanced distribution in the dataset, we split it into subsets by domains or models. Then we trained a threshold-based classifier for each feature across different subsets and analyzed their performance.

Interestingly, some features consistently exhibit high classification quality across multiple domains, which we refer to as general features. In contrast, other features are more specialized, performing well only within specific domains or detecting generations of a particular subset of models, highlighting their domain- or model-specific nature. Examples of these features and their performance are shown in Figure \ref{fig:domains_heatmap}.

Some general features (e.g., 3608 and 4645 in layer 16) appear universal across domains and models. To demonstrate this, we compare the best feature for detecting each generator to these universal features (Figure \ref{fig:models_bars}). The graph shows that for older models (e.g., flan, t0), universal feature performance drops below random, while the \textit{opt} family is the most "universal." This suggests distinct characteristics among model classes: older/weaker models (flan, t0), more advanced LLMs (opt, bloom, gpt\_j, gpt\_neo), and modern families (GPT-3.5+, LLaMa, Gemma). The next section explores these differences further.

\noindent \textbf{Robust Feature Analysis.} Building on \citet{kuznetsov-etal-2024-robust}, we evaluate the classifier for the presence of harmful superficial features and those vulnerable to different types of attacks on artificial text classifiers, using the RAID dataset. Details on feature extraction can be found in Appendix \ref{sec:sensitive_features}. Our analysis shows that features most susceptible to attacks and shallow text properties overlap minimally with those identified as important by XGBoost. Specifically, features 8689 (detecting the GPT3.5+ family) and 14919 (detecting the Bloom family) are very sensitive to sentence length, while other distractions have limited impact on important features.

\section{Important Features Interpretation}

In this section, we discuss the insights from analyzing our feature interpretations (see detailes and examples in Appendix~\ref{appendix:feature_interpretations}), starting with the most robust features: 3608, 4645, 6587, 8264, and 14161. Their performance in the ATD task across various domains and models is shown in Figures~\ref{fig:domains_heatmap} and \ref{fig:models_heatmap}.

Strong activations of these features correlate with common LLM-generated text characteristics, such as excessive complexity (3608), assertive claims (4645), wordy introductions (6587), repetition (8264), and formality (14161). These features perform well on GPT3.5+ and other modern LLMs like LLaMa and Gemma, especially for domains like finance, medicine, and Wiki-CSAi. However, texts from arXiv are less distinguishable, suggesting GPT models mimic scientific writing more closely.

Feature 8264 stands out with near-perfect performance for GPT3.5+, controlling the conciseness vs. repetition of concepts. Older models lack this feature, leading to lower detectability.

Domain-specific features include overcomplicated syntax (arXiv, feature 12390), excessive details (finance, feature 6513), speculative links (Reddit, feature 4560), and hallucinated facts (Wikipedia, feature 4773). Improper tone (medicine, feature 14953) also signals machine-generated texts.

The most challenging domains for detection are Outfox (essays) and Yelp (reviews), where models mimic human-like writing. This suggests that general “overcomplexity” features may not be effective when models are instructed to avoid such traits.

\section{Conclusion}

Our analysis shows that modern LLMs often generate easily detectable text due to specific writing styles, such as long-winded introductions, excessive synonym substitution, and repetition. However, adversaries can bypass these features by using less formal, more personalized prompts, like student essays, leading to more human-like outputs. 

Unlike previous approaches, we perform a multi-faceted analysis of features for Artificial Text Detection (ATD). We select key features, examine their behavior across domains and generators, and interpret them both through extreme values (manual) and medium shifts (steering + LLM interpretation). This approach provides deeper insights into feature meanings. For example, our interpretation of feature 3608 contrasts with Neuropedia's narrow view, which links it to "tokens associated with mathematical expressions." Similarly, feature 4645, described by Neuropedia as related to "keywords on diabetes," is more broadly relevant in our analysis.

We conclude that Sparse Autoencoder-based analysis of ATD datasets is a valuable tool for understanding text generators, detectors, and how detectors generalize to new setups. Our findings highlight that detecting AI-generated text is easy with a default prompt but becomes difficult when prompt style changes, a crucial consideration for ATD developers.

\section{Limitations}

Artificial text detection (ATD) is a highly complex and evolving task. With new LLMs emerging almost every month, it is difficult to predict how our method will perform on future artificial text generators. Additionally, novel attack strategies continue to appear, and our approach covers only a subset of them. Besides, some of SAE features we studied remain challenging to interpret, as not all exhibit clear semantic meaning.

Finally, in this short paper, we used a single Sparse Autoencoder (SAE) on the residual stream of Gemma 2-2B. Exploring different SAEs on other LLMs could reveal new features and offer additional insights into artificial text detection. We leave this for future work.

\bibliography{custom}

\begin{thebibliography}{48}
\providecommand{\natexlab}[1]{#1}

\bibitem[{Abdali et~al.(2024)Abdali, Anarfi, Barberan, and He}]{abdali2024decoding}
Sara Abdali, Richard Anarfi, CJ~Barberan, and Jia He. 2024.
\newblock Decoding the ai pen: Techniques and challenges in detecting ai-generated text.
\newblock In \emph{Proceedings of the 30th ACM SIGKDD Conference on Knowledge Discovery and Data Mining}, pages 6428--6436.

\bibitem[{Black et~al.(2022)Black, Biderman, Hallahan, Anthony, Gao, Golding, He, Leahy, McDonell, Phang, Pieler, Prashanth, Purohit, Reynolds, Tow, Wang, and Weinbach}]{black-etal-2022-gpt}
Sidney Black, Stella Biderman, Eric Hallahan, Quentin Anthony, Leo Gao, Laurence Golding, Horace He, Connor Leahy, Kyle McDonell, Jason Phang, Michael Pieler, Usvsn~Sai Prashanth, Shivanshu Purohit, Laria Reynolds, Jonathan Tow, Ben Wang, and Samuel Weinbach. 2022.
\newblock \href {https://doi.org/10.18653/v1/2022.bigscience-1.9} {{GPT}-{N}eo{X}-20{B}: An open-source autoregressive language model}.
\newblock In \emph{Proceedings of BigScience Episode {\#}5 -- Workshop on Challenges {\&} Perspectives in Creating Large Language Models}, pages 95--136, virtual+Dublin. Association for Computational Linguistics.

\bibitem[{Bricken et~al.(2023)Bricken, Templeton, Batson, Chen, Jermyn, Conerly, Turner, Anil, Denison, Askell, Lasenby, Wu, Kravec, Maxwell, Joseph, Hatfield-Dodds, Tamkin, Nguyen, McLean, Burke, Hume, Carter, Henighan, and Olah}]{Bricken2023}
Trenton Bricken, Adly Templeton, Joshua Batson, Brian Chen, Adam Jermyn, Tom Conerly, Nick Turner, Cem Anil, Carson Denison, Amanda Askell, Robert Lasenby, Yifan Wu, Shauna Kravec, Tim Maxwell, Nicholas Joseph, Zac Hatfield-Dodds, Alex Tamkin, Karina Nguyen, Brayden McLean, Josiah Burke, Tristan Hume, Shan Carter, Tom Henighan, and Christopher Olah. 2023.
\newblock Towards monosemanticity: Decomposing language models with dictionary learning.
\newblock \url{https://transformer-circuits.pub/2023/monosemantic-features/index.html}.
\newblock Transformer Circuits Thread.

\bibitem[{Cai and Cui(2023)}]{cai2023evadechatgptdetectorssingle}
Shuyang Cai and Wanyun Cui. 2023.
\newblock \href {https://arxiv.org/abs/2307.02599} {Evade chatgpt detectors via a single space}.
\newblock \emph{Preprint}, arXiv:2307.02599.

\bibitem[{Chakraborty et~al.(2023)Chakraborty, Bedi, Zhu, An, Manocha, and Huang}]{Chakraborty2023Possibilities}
Souradip Chakraborty, A.~S. Bedi, Sicheng Zhu, Bang An, Dinesh Manocha, and Furong Huang. 2023.
\newblock \href {https://arxiv.org/abs/2304.04736} {On the possibilities of {AI}-generated text detection}.
\newblock \emph{arXiv preprint arXiv:2304.04736}.

\bibitem[{Chen et~al.(2023)Chen, Kang, Zhai, Li, Singh, and Raj}]{chen2023gpt}
Yutian Chen, Hao Kang, Vivian Zhai, Liangze Li, Rita Singh, and Bhiksha Raj. 2023.
\newblock Gpt-sentinel: Distinguishing human and chatgpt generated content.
\newblock \emph{arXiv preprint arXiv:2305.07969}.

\bibitem[{Cunningham et~al.(2023)Cunningham, Ewart, Riggs, Huben, and Sharkey}]{Cunningham2023}
Hoagy Cunningham, Aidan Ewart, Logan~R. Riggs, Robert Huben, and Lee Sharkey. 2023.
\newblock Sparse autoencoders find highly interpretable features in language models.
\newblock \emph{arXiv preprint arXiv:2309.08600}.

\bibitem[{Dugan et~al.(2024)Dugan, Hwang, Trhl{\'i}k, Zhu, Ludan, Xu, Ippolito, and Callison-Burch}]{dugan-etal-2024-raid}
Liam Dugan, Alyssa Hwang, Filip Trhl{\'i}k, Andrew Zhu, Josh~Magnus Ludan, Hainiu Xu, Daphne Ippolito, and Chris Callison-Burch. 2024.
\newblock \href {https://doi.org/10.18653/v1/2024.acl-long.674} {{RAID}: A shared benchmark for robust evaluation of machine-generated text detectors}.
\newblock In \emph{Proceedings of the 62nd Annual Meeting of the Association for Computational Linguistics (Volume 1: Long Papers)}, pages 12463--12492, Bangkok, Thailand. Association for Computational Linguistics.

\bibitem[{Elhage et~al.(2022{\natexlab{a}})Elhage, Hume, Olsson, Schiefer, Henighan, Kravec, Hatfield-Dodds, Lasenby, Drain, Chen, Grosse, McCandlish, Kaplan, Amodei, Wattenberg, and Olah}]{elhage2022superposition}
Nelson Elhage, Tristan Hume, Catherine Olsson, Nicholas Schiefer, Tom Henighan, Shauna Kravec, Zac Hatfield-Dodds, Robert Lasenby, Dawn Drain, Carol Chen, Roger Grosse, Sam McCandlish, Jared Kaplan, Dario Amodei, Martin Wattenberg, and Christopher Olah. 2022{\natexlab{a}}.
\newblock Toy models of superposition.
\newblock \emph{Transformer Circuits Thread}.

\bibitem[{Elhage et~al.(2022{\natexlab{b}})Elhage, Hume, Olsson, Schiefer, Henighan, Kravec, Hatfield-Dodds, Lasenby, Drain, Chen et~al.}]{elhage2022toy}
Nelson Elhage, Tristan Hume, Catherine Olsson, Nicholas Schiefer, Tom Henighan, Shauna Kravec, Zac Hatfield-Dodds, Robert Lasenby, Dawn Drain, Carol Chen, et~al. 2022{\natexlab{b}}.
\newblock Toy models of superposition.
\newblock \emph{arXiv preprint arXiv:2209.10652}.

\bibitem[{Elhage et~al.(2023)Elhage, Lasenby, and Olah}]{elhage2023privileged}
Nelson Elhage, Robert Lasenby, and Christopher Olah. 2023.
\newblock Privileged bases in the transformer residual stream, 2023.
\newblock \emph{URL https://transformer-circuits. pub/2023/privilegedbasis/index. html Accessed: 2024-01-14}.

\bibitem[{Gao et~al.(2023)Gao, Dupré~la Tour, Tillman, Goh, Troll, Radford, Sutskever, Leike, and Wu}]{Gao2023}
Leo Gao, Tom Dupré~la Tour, Henk Tillman, Gabriel Goh, Rajan Troll, Alec Radford, Ilya Sutskever, Jan Leike, and Jeffrey Wu. 2023.
\newblock Scaling and evaluating sparse autoencoders.
\newblock OpenAI Technical Report. \url{https://cdn.openai.com/papers/sparse-autoencoders.pdf}.

\bibitem[{Gehrmann et~al.(2019)Gehrmann, Strobelt, and Rush}]{Gehrmann2019GLTR}
Sebastian Gehrmann, Hendrik Strobelt, and Alexander~M. Rush. 2019.
\newblock \href {https://doi.org/10.18653/v1/P19-3019} {{GLTR}: Statistical detection and visualization of generated text}.
\newblock In \emph{Proceedings of the 57th Annual Meeting of the Association for Computational Linguistics: System Demonstrations}, pages 111--116.

\bibitem[{Grattafiori et~al.(2024)}]{grattafiori2024llama3herdmodels}
Aaron Grattafiori et~al. 2024.
\newblock \href {https://arxiv.org/abs/2407.21783} {The llama 3 herd of models}.
\newblock \emph{Preprint}, arXiv:2407.21783.

\bibitem[{Gritsai et~al.(2025)Gritsai, Voznyuk, Khabutdinov, and Grabovoy}]{gritsai-etal-2025-advacheck}
German Gritsai, Anastasia Voznyuk, Ildar Khabutdinov, and Andrey Grabovoy. 2025.
\newblock \href {https://aclanthology.org/2025.genaidetect-1.26/} {Advacheck at {G}en{AI} detection task 1: {AI} detection powered by domain-aware multi-tasking}.
\newblock In \emph{Proceedings of the 1stWorkshop on GenAI Content Detection (GenAIDetect)}, pages 236--243, Abu Dhabi, UAE. International Conference on Computational Linguistics.

\bibitem[{Guo et~al.(2023)Guo, Zhang, Wang, Jiang, Nie, Ding, Yue, and Wu}]{guo2023close}
Biyang Guo, Xin Zhang, Ziyuan Wang, Minqi Jiang, Jinran Nie, Yuxuan Ding, Jianwei Yue, and Yupeng Wu. 2023.
\newblock How close is chatgpt to human experts? comparison corpus, evaluation, and detection.
\newblock \emph{arXiv preprint arXiv:2301.07597}.

\bibitem[{Huben et~al.(2023)Huben, Cunningham, Smith, Ewart, and Sharkey}]{huben2023sparse}
Robert Huben, Hoagy Cunningham, Logan~Riggs Smith, Aidan Ewart, and Lee Sharkey. 2023.
\newblock Sparse autoencoders find highly interpretable features in language models.
\newblock In \emph{The Twelfth International Conference on Learning Representations}.

\bibitem[{Jiang et~al.(2023)Jiang, Sablayrolles, Mensch, Bamford, Chaplot, de~las Casas, Bressand, Lengyel, Lample, Saulnier, Lavaud, Lachaux, Stock, Scao, Lavril, Wang, Lacroix, and Sayed}]{jiang2023mistral}
Albert~Q. Jiang, Alexandre Sablayrolles, Arthur Mensch, Chris Bamford, Devendra~Singh Chaplot, Diego de~las Casas, Florian Bressand, Gianna Lengyel, Guillaume Lample, Lucile Saulnier, Lélio~Renard Lavaud, Marie-Anne Lachaux, Pierre Stock, Teven~Le Scao, Thibaut Lavril, Thomas Wang, Timothée Lacroix, and William~El Sayed. 2023.
\newblock \href {https://arxiv.org/abs/2310.06825} {Mistral 7b}.
\newblock \emph{Preprint}, arXiv:2310.06825.

\bibitem[{Jiang et~al.(2024)Jiang, Sablayrolles, Roux, Mensch, Savary, Bamford, Chaplot, de~las Casas, Hanna, Bressand, Lengyel, Bour, Lample, Lavaud, Saulnier, Lachaux, Stock, Subramanian, Yang, Antoniak, Scao, Gervet, Lavril, Wang, Lacroix, and Sayed}]{jiang2024mixtralexperts}
Albert~Q. Jiang, Alexandre Sablayrolles, Antoine Roux, Arthur Mensch, Blanche Savary, Chris Bamford, Devendra~Singh Chaplot, Diego de~las Casas, Emma~Bou Hanna, Florian Bressand, Gianna Lengyel, Guillaume Bour, Guillaume Lample, Lélio~Renard Lavaud, Lucile Saulnier, Marie-Anne Lachaux, Pierre Stock, Sandeep Subramanian, Sophia Yang, Szymon Antoniak, Teven~Le Scao, Théophile Gervet, Thibaut Lavril, Thomas Wang, Timothée Lacroix, and William~El Sayed. 2024.
\newblock \href {https://arxiv.org/abs/2401.04088} {Mixtral of experts}.
\newblock \emph{Preprint}, arXiv:2401.04088.

\bibitem[{Krishna et~al.(2023)Krishna, Song, Karpinska, Wieting, and Iyyer}]{krishna2023paraphrasing}
Kalpesh Krishna, Yixiao Song, Marzena Karpinska, John Wieting, and Mohit Iyyer. 2023.
\newblock \href {https://arxiv.org/abs/2303.13408} {Paraphrasing evades detectors of ai-generated text, but retrieval is an effective defense}.
\newblock \emph{Preprint}, arXiv:2303.13408.

\bibitem[{Kushnareva et~al.(2024)Kushnareva, Gaintseva, Magai, Barannikov, Abulkhanov, Kuznetsov, Tulchinskii, Piontkovskaya, and Nikolenko}]{kushnareva2024aigeneratedtextboundarydetection}
Laida Kushnareva, Tatiana Gaintseva, German Magai, Serguei Barannikov, Dmitry Abulkhanov, Kristian Kuznetsov, Eduard Tulchinskii, Irina Piontkovskaya, and Sergey Nikolenko. 2024.
\newblock \href {https://arxiv.org/abs/2311.08349} {Ai-generated text boundary detection with roft}.
\newblock \emph{Preprint}, arXiv:2311.08349.

\bibitem[{Kuznetsov et~al.(2024)Kuznetsov, Tulchinskii, Kushnareva, Magai, Barannikov, Nikolenko, and Piontkovskaya}]{kuznetsov-etal-2024-robust}
Kristian Kuznetsov, Eduard Tulchinskii, Laida Kushnareva, German Magai, Serguei Barannikov, Sergey Nikolenko, and Irina Piontkovskaya. 2024.
\newblock \href {https://doi.org/10.18653/v1/2024.findings-emnlp.992} {Robust {AI}-generated text detection by restricted embeddings}.
\newblock In \emph{Findings of the Association for Computational Linguistics: EMNLP 2024}, pages 17036--17055, Miami, Florida, USA. Association for Computational Linguistics.

\bibitem[{Li et~al.(2023)Li, Li, Cui, Bi, Wang, Yang, Shi, and Zhang}]{Li2023DeepfakeWild}
Yafu Li, Qintong Li, Leyang Cui, Wei Bi, Longyue Wang, Linyi Yang, Shuming Shi, and Yue Zhang. 2023.
\newblock \href {https://arxiv.org/abs/2305.13242} {Deepfake text detection in the wild}.
\newblock \emph{arXiv preprint arXiv:2305.13242}.

\bibitem[{Lieberum et~al.(2024)Lieberum, Rajamanoharan, Conmy, Smith, Sonnerat, Varma, Kramar, Dragan, Shah, and Nanda}]{lieberum-etal-2024-gemma}
Tom Lieberum, Senthooran Rajamanoharan, Arthur Conmy, Lewis Smith, Nicolas Sonnerat, Vikrant Varma, Janos Kramar, Anca Dragan, Rohin Shah, and Neel Nanda. 2024.
\newblock \href {https://doi.org/10.18653/v1/2024.blackboxnlp-1.19} {Gemma scope: Open sparse autoencoders everywhere all at once on gemma 2}.
\newblock In \emph{Proceedings of the 7th BlackboxNLP Workshop: Analyzing and Interpreting Neural Networks for NLP}, pages 278--300, Miami, Florida, US. Association for Computational Linguistics.

\bibitem[{Makelov et~al.(2024)Makelov, Lange, and Nanda}]{makelov2024towards}
Aleksandar Makelov, George Lange, and Neel Nanda. 2024.
\newblock Towards principled evaluations of sparse autoencoders for interpretability and control.
\newblock \emph{arXiv preprint arXiv:2405.08366}.

\bibitem[{Mitchell et~al.(2023)Mitchell, Lee, Khazatsky, Manning, and Finn}]{Mitchell2023DetectGPT}
Eric Mitchell, Yoonho Lee, Alexander Khazatsky, Christopher~D. Manning, and Chelsea Finn. 2023.
\newblock {DetectGPT}: Zero-shot machine-generated text detection using probability curvature.
\newblock \emph{arXiv preprint arXiv:2301.11305}.

\bibitem[{Muennighoff et~al.(2023)Muennighoff, Wang, Sutawika, Roberts, Biderman, Scao, Bari, Shen, Yong, Schoelkopf, Tang, Radev, Aji, Almubarak, Albanie, Alyafeai, Webson, Raff, and Raffel}]{muennighoff2023crosslingualgeneralizationmultitaskfinetuning}
Niklas Muennighoff, Thomas Wang, Lintang Sutawika, Adam Roberts, Stella Biderman, Teven~Le Scao, M~Saiful Bari, Sheng Shen, Zheng-Xin Yong, Hailey Schoelkopf, Xiangru Tang, Dragomir Radev, Alham~Fikri Aji, Khalid Almubarak, Samuel Albanie, Zaid Alyafeai, Albert Webson, Edward Raff, and Colin Raffel. 2023.
\newblock \href {https://arxiv.org/abs/2211.01786} {Crosslingual generalization through multitask finetuning}.
\newblock \emph{Preprint}, arXiv:2211.01786.

\bibitem[{Olah et~al.(2020)Olah, Cammarata, Schubert, Goh, Petrov, and Carter}]{olah2020zoom}
Chris Olah, Nick Cammarata, Ludwig Schubert, Gabriel Goh, Michael Petrov, and Shan Carter. 2020.
\newblock \href {https://doi.org/10.23915/distill.00024.001} {Zoom in: An introduction to circuits}.
\newblock \emph{Distill}.
\newblock Https://distill.pub/2020/circuits/zoom-in.

\bibitem[{OpenAI(2024{\natexlab{a}})}]{openai2024gpt4technicalreport}
OpenAI. 2024{\natexlab{a}}.
\newblock \href {https://arxiv.org/abs/2303.08774} {Gpt-4 technical report}.
\newblock \emph{Preprint}, arXiv:2303.08774.

\bibitem[{OpenAI(2024{\natexlab{b}})}]{openai2024gpt4ocard}
OpenAI. 2024{\natexlab{b}}.
\newblock \href {https://arxiv.org/abs/2410.21276} {Gpt-4o system card}.
\newblock \emph{Preprint}, arXiv:2410.21276.

\bibitem[{Radford et~al.(2019)Radford, Wu, Child, Luan, Amodei, Sutskever et~al.}]{radford2019language}
Alec Radford, Jeffrey Wu, Rewon Child, David Luan, Dario Amodei, Ilya Sutskever, et~al. 2019.
\newblock Language models are unsupervised multitask learners.
\newblock \emph{OpenAI blog}, 1(8):9.

\bibitem[{Sanh et~al.(2022)Sanh, Webson, Raffel, Bach, Sutawika, Alyafeai, Chaffin, Stiegler, Scao, Raja, Dey, Bari, Xu, Thakker, Sharma, Szczechla, Kim, Chhablani, Nayak, Datta, Chang, Jiang, Wang, Manica, Shen, Yong, Pandey, Bawden, Wang, Neeraj, Rozen, Sharma, Santilli, Fevry, Fries, Teehan, Bers, Biderman, Gao, Wolf, and Rush}]{sanh2022multitaskpromptedtrainingenables}
Victor Sanh, Albert Webson, Colin Raffel, Stephen~H. Bach, Lintang Sutawika, Zaid Alyafeai, Antoine Chaffin, Arnaud Stiegler, Teven~Le Scao, Arun Raja, Manan Dey, M~Saiful Bari, Canwen Xu, Urmish Thakker, Shanya~Sharma Sharma, Eliza Szczechla, Taewoon Kim, Gunjan Chhablani, Nihal Nayak, Debajyoti Datta, Jonathan Chang, Mike Tian-Jian Jiang, Han Wang, Matteo Manica, Sheng Shen, Zheng~Xin Yong, Harshit Pandey, Rachel Bawden, Thomas Wang, Trishala Neeraj, Jos Rozen, Abheesht Sharma, Andrea Santilli, Thibault Fevry, Jason~Alan Fries, Ryan Teehan, Tali Bers, Stella Biderman, Leo Gao, Thomas Wolf, and Alexander~M. Rush. 2022.
\newblock \href {https://arxiv.org/abs/2110.08207} {Multitask prompted training enables zero-shot task generalization}.
\newblock \emph{Preprint}, arXiv:2110.08207.

\bibitem[{Schulman et~al.(2022)}]{chatgptpost}
John Schulman et~al. 2022.
\newblock \href {https://openai.com/index/chatgpt/} {Introducing chatgpt}.

\bibitem[{Sharkey et~al.(2023)Sharkey, Braun, and Millidge}]{shakley2023superposition}
Lee Sharkey, Dan Braun, and Beren Millidge. 2023.
\newblock Taking features out of superposition with sparse autoencoders, 2023.
\newblock \emph{URL https://www.lesswrong. com/posts/z6QQJbtpkEAX3Aojj/interim-research-report-taking-features-out-of-superposition. Accessed: 2024-01-14.}

\bibitem[{Solaiman et~al.(2019)Solaiman, Brundage, Clark, Askell, Herbert-Voss, Wu, Radford, and Wang}]{Solaiman2019ReleaseLM}
Irene Solaiman, Miles Brundage, Jack Clark, Amanda Askell, Ariel Herbert-Voss, Jeff Wu, Alec Radford, and Jasmine Wang. 2019.
\newblock \href {https://arxiv.org/abs/1908.09203} {Release strategies and the social impacts of language models}.
\newblock \emph{arXiv preprint arXiv:1908.09203}.

\bibitem[{Team(2024{\natexlab{a}})}]{gemmateam2024gemma2improvingopen}
Gemma Team. 2024{\natexlab{a}}.
\newblock \href {https://arxiv.org/abs/2408.00118} {Gemma 2: Improving open language models at a practical size}.
\newblock \emph{Preprint}, arXiv:2408.00118.

\bibitem[{Team(2024{\natexlab{b}})}]{gemmateam2024gemmaopenmodelsbased}
Gemma Team. 2024{\natexlab{b}}.
\newblock \href {https://arxiv.org/abs/2403.08295} {Gemma: Open models based on gemini research and technology}.
\newblock \emph{Preprint}, arXiv:2403.08295.

\bibitem[{Touvron et~al.(2023)Touvron, Lavril, Izacard, Martinet, Lachaux, Lacroix, Rozière, Goyal, Hambro, Azhar, Rodriguez, Joulin, Grave, and Lample}]{touvron2023llamaopenefficientfoundation}
Hugo Touvron, Thibaut Lavril, Gautier Izacard, Xavier Martinet, Marie-Anne Lachaux, Timothée Lacroix, Baptiste Rozière, Naman Goyal, Eric Hambro, Faisal Azhar, Aurelien Rodriguez, Armand Joulin, Edouard Grave, and Guillaume Lample. 2023.
\newblock \href {https://arxiv.org/abs/2302.13971} {Llama: Open and efficient foundation language models}.
\newblock \emph{Preprint}, arXiv:2302.13971.

\bibitem[{Tulchinskii et~al.(2023)Tulchinskii, Kuznetsov, Kushnareva, Cherniavskii, Nikolenko, Burnaev, Barannikov, and Piontkovskaya}]{tulchinskii2023intrinsic}
Eduard Tulchinskii, Kristian Kuznetsov, Laida Kushnareva, Daniil Cherniavskii, Sergey Nikolenko, Evgeny Burnaev, Serguei Barannikov, and Irina Piontkovskaya. 2023.
\newblock Intrinsic dimension estimation for robust detection of ai-generated texts.
\newblock \emph{Advances in Neural Information Processing Systems}, 36:39257--39276.

\bibitem[{Uchendu et~al.(2021)Uchendu, Ma, Le, Zhang, and Lee}]{Uchendu2021TuringBench}
Adaku Uchendu, Zeyu Ma, Thai Le, Rui Zhang, and Dongwon Lee. 2021.
\newblock \href {https://doi.org/10.18653/v1/2021.findings-emnlp.172} {{TURINGBENCH}: A benchmark environment for turing test in the age of neural text generation}.
\newblock In \emph{Findings of the Association for Computational Linguistics: EMNLP 2021}, pages 2001--2017.

\bibitem[{Wang and Komatsuzaki(2021)}]{gptj}
Ben Wang and Aran Komatsuzaki. 2021.
\newblock {GPT-J-6B: A 6 Billion Parameter Autoregressive Language Model}.
\newblock \url{https://github.com/kingoflolz/mesh-transformer-jax}.

\bibitem[{Wang et~al.(2025)Wang, Shelmanov, Mansurov, Tsvigun, Mikhailov, Xing, Xie, Geng, Puccetti, Artemova, Su, Ta, Abassy, Elozeiri, El~Etter, Goloburda, Mahmoud, Tomar, Laiyk, Mohammed~Afzal, Koike, Kaneko, Aji, Habash, Gurevych, and Nakov}]{wang-etal-2025-genai}
Yuxia Wang, Artem Shelmanov, Jonibek Mansurov, Akim Tsvigun, Vladislav Mikhailov, Rui Xing, Zhuohan Xie, Jiahui Geng, Giovanni Puccetti, Ekaterina Artemova, Jinyan Su, Minh~Ngoc Ta, Mervat Abassy, Kareem~Ashraf Elozeiri, Saad El Dine~Ahmed El~Etter, Maiya Goloburda, Tarek Mahmoud, Raj~Vardhan Tomar, Nurkhan Laiyk, Osama Mohammed~Afzal, Ryuto Koike, Masahiro Kaneko, Alham~Fikri Aji, Nizar Habash, Iryna Gurevych, and Preslav Nakov. 2025.
\newblock \href {https://aclanthology.org/2025.genaidetect-1.27/} {{G}en{AI} content detection task 1: {E}nglish and multilingual machine-generated text detection: {AI} vs. human}.
\newblock In \emph{Proceedings of the 1stWorkshop on GenAI Content Detection (GenAIDetect)}, pages 244--261, Abu Dhabi, UAE. International Conference on Computational Linguistics.

\bibitem[{Xue et~al.(2021)Xue, Constant, Roberts, Kale, Al-Rfou, Siddhant, Barua, and Raffel}]{xue2021mt5massivelymultilingualpretrained}
Linting Xue, Noah Constant, Adam Roberts, Mihir Kale, Rami Al-Rfou, Aditya Siddhant, Aditya Barua, and Colin Raffel. 2021.
\newblock \href {https://arxiv.org/abs/2010.11934} {mt5: A massively multilingual pre-trained text-to-text transformer}.
\newblock \emph{Preprint}, arXiv:2010.11934.

\bibitem[{Yang et~al.(2024)Yang, Pan, Zhao, Chen, Petzold, Wang, and Cheng}]{yang-etal-2024-survey}
Xianjun Yang, Liangming Pan, Xuandong Zhao, Haifeng Chen, Linda~Ruth Petzold, William~Yang Wang, and Wei Cheng. 2024.
\newblock \href {https://doi.org/10.18653/v1/2024.findings-emnlp.572} {A survey on detection of {LLM}s-generated content}.
\newblock In \emph{Findings of the Association for Computational Linguistics: EMNLP 2024}, pages 9786--9805, Miami, Florida, USA. Association for Computational Linguistics.

\bibitem[{Yu et~al.(2023)Yu, Qi, Chen, Chen, Yang, Zhu, Zhang, and Yu}]{yu2023gpt}
Xiao Yu, Yuang Qi, Kejiang Chen, Guoqiang Chen, Xi~Yang, Pengyuan Zhu, Weiming Zhang, and Nenghai Yu. 2023.
\newblock Gpt paternity test: Gpt generated text detection with gpt genetic inheritance.
\newblock \emph{CoRR}.

\bibitem[{Zellers et~al.(2019)Zellers, Holtzman, Rashkin, Bisk, Farhadi, Roesner, and Choi}]{Zellers2019Grover}
Rowan Zellers, Ari Holtzman, Hannah Rashkin, Yonatan Bisk, Ali Farhadi, Franziska Roesner, and Yejin Choi. 2019.
\newblock \href {https://papers.nips.cc/paper/2019/hash/3e9f0fc9b2f89e043bc6233994dfcf76-Abstract.html} {Defending against neural fake news}.
\newblock \emph{Advances in Neural Information Processing Systems (NeurIPS)}, 32.

\bibitem[{Zeng et~al.(2023)Zeng, Liu, Du, Wang, Lai, Ding, Yang, Xu, Zheng, Xia, Tam, Ma, Xue, Zhai, Chen, Zhang, Dong, and Tang}]{zeng2023glm130bopenbilingualpretrained}
Aohan Zeng, Xiao Liu, Zhengxiao Du, Zihan Wang, Hanyu Lai, Ming Ding, Zhuoyi Yang, Yifan Xu, Wendi Zheng, Xiao Xia, Weng~Lam Tam, Zixuan Ma, Yufei Xue, Jidong Zhai, Wenguang Chen, Peng Zhang, Yuxiao Dong, and Jie Tang. 2023.
\newblock \href {https://arxiv.org/abs/2210.02414} {Glm-130b: An open bilingual pre-trained model}.
\newblock \emph{Preprint}, arXiv:2210.02414.

\bibitem[{Zhang et~al.(2022)Zhang, Roller, Goyal, Artetxe, Chen, Chen, Dewan, Diab, Li, Lin, Mihaylov, Ott, Shleifer, Shuster, Simig, Koura, Sridhar, Wang, and Zettlemoyer}]{zhang2022optopenpretrainedtransformer}
Susan Zhang, Stephen Roller, Naman Goyal, Mikel Artetxe, Moya Chen, Shuohui Chen, Christopher Dewan, Mona Diab, Xian Li, Xi~Victoria Lin, Todor Mihaylov, Myle Ott, Sam Shleifer, Kurt Shuster, Daniel Simig, Punit~Singh Koura, Anjali Sridhar, Tianlu Wang, and Luke Zettlemoyer. 2022.
\newblock \href {https://arxiv.org/abs/2205.01068} {Opt: Open pre-trained transformer language models}.
\newblock \emph{Preprint}, arXiv:2205.01068.

\end{thebibliography}

\newpage

\appendix

\section{Related Work}
\label{sec:related-work}

\noindent \textbf{Machine-Generated Text Detection.} Detection systems for distinguishing human and AI-generated text follow two main approaches: Training-Based and Zero-Shot methods. Training-based approaches fine-tune Transformer models on labeled datasets for strong in-domain performance \cite{chen2023gpt, Li2023DeepfakeWild, yu2023gpt}. In contrast, zero-shot methods analyze statistical patterns without supervised fine-tuning, like  token likelihoods, probability curvature or intrinsic dimension \cite{Gehrmann2019GLTR, Mitchell2023DetectGPT, tulchinskii2023intrinsic}. 

However, the challenge of making AI-generated text more interpretable for humans has only been addressed by a limited number of approaches, either through manual analysis \cite{guo2023close} or only partially investingating the dependencies \cite{kuznetsov-etal-2024-robust}.

\noindent \textbf{Sparse Autoencoders and Interpretability.} LLM interpretability is especially challenging due to polysemanticity, where a single neuron encodes multiple unrelated concepts \cite{elhage2022superposition, elhage2023privileged}. Sparse Autoencoders (SAEs) were proposed to help isolating more interpretable latent dimensions \cite{shakley2023superposition}. Unlike standard autoencoders, SAEs introduce a penalty (e.g. $L_1$ regularization) to ensure that only a small subset of neurons is active per input, resulting in highly interpretable features \cite{Cunningham2023}.

Recent approaches use large language models or heuristics to automate hypothesis generation and refinement \cite{Bricken2023, Cunningham2023, Gao2023}. For example, \cite{Bricken2023} employ GPT-4 to label sparse dimensions based on top-activating tokens, while \cite{Cunningham2023} use heuristic methods like measuring overlap with linguistic categories to infer dimension meanings. In our work we employ both manual and automatic interpretation to ensure unbiasedness of our approach.

\noindent \textbf{Datasets and Benchmarks.} AI text detection includes many datasets, starting with GPT-2 Output \cite{Solaiman2019ReleaseLM} and Grover \cite{Zellers2019Grover}, as well as TuringBench \cite{Uchendu2021TuringBench}, which unifies 19 models for cross-evaluation. Additionally, domain-specific corpora and “in-the-wild” tests, such as \cite{Chakraborty2023Possibilities}, become useful for enhancing model robustness.

\section{COLING dataset: additional details}
\label{sec:generation_examples}

The COLING dataset contains generations of the models from the following families: a) LLaMA, 7 - 65B \cite{touvron2023llamaopenefficientfoundation}; b) LLaMA 3, 8 and 70B \cite{grattafiori2024llama3herdmodels}; c) GLM, 130B \cite{zeng2023glm130bopenbilingualpretrained}; d) Bloomz and Bloom 7B \cite{muennighoff2023crosslingualgeneralizationmultitaskfinetuning}; e) cohere\footnote{\url{https://docs.cohere.com/docs/models}}; f) GPT 3.5 series, including davinci 001-003 model\footnote{\url{https://platform.openai.com/docs/models}} and gpt-3.5-turbo \cite{chatgptpost}; g) GPT-4 \cite{openai2024gpt4technicalreport} and GPT-4-o \cite{openai2024gpt4ocard}; h) a line of models, based on T5 \cite{xue2021mt5massivelymultilingualpretrained} and T0 \cite{sanh2022multitaskpromptedtrainingenables}; i) Gemma, 7B \cite{gemmateam2024gemmaopenmodelsbased} and Gemma 2, 9B \cite{gemmateam2024gemma2improvingopen}; j) GPT-J, 6B \cite{gptj} and GPT-Neo-X, 20B \cite{black-etal-2022-gpt}; k) Mixtral, 8 x 7B \cite{jiang2024mixtralexperts}; l) OPT, 125M - 30B \cite{zhang2022optopenpretrainedtransformer}.

After analyzing the dataset manually, we identified that some samples contain anomalous punctuation. Figures~\ref{fig:anomalous_samples_gpt} and~\ref{fig:anomalous_samples_human} display several fragments of such samples. For comparison, Figures~\ref{fig:normal_samples_gpt} and~\ref{fig:normal_samples_human} show samples from the same models (or human texts) without these anomalies. We hypothesize that this inconsistency arises from the COLING dataset being composed of multiple datasets created by different authors.

Previous research works have shown that spurious features related to the text length \cite{kushnareva2024aigeneratedtextboundarydetection} and formatting \cite{dugan-etal-2024-raid} significantly affect artificial text detection. Moreover, \citet{cai2023evadechatgptdetectorssingle} found that sometimes adding even a single space before the comma may confuse detectors. Thus, we find it important to analyze the peculiarities of the dataset we use and investigate whether the features we examine truly reflect inherent properties of the generated texts or are simply influenced by superficial traits.

\begin{figure*}[t]
  \includegraphics[width=0.99\linewidth]{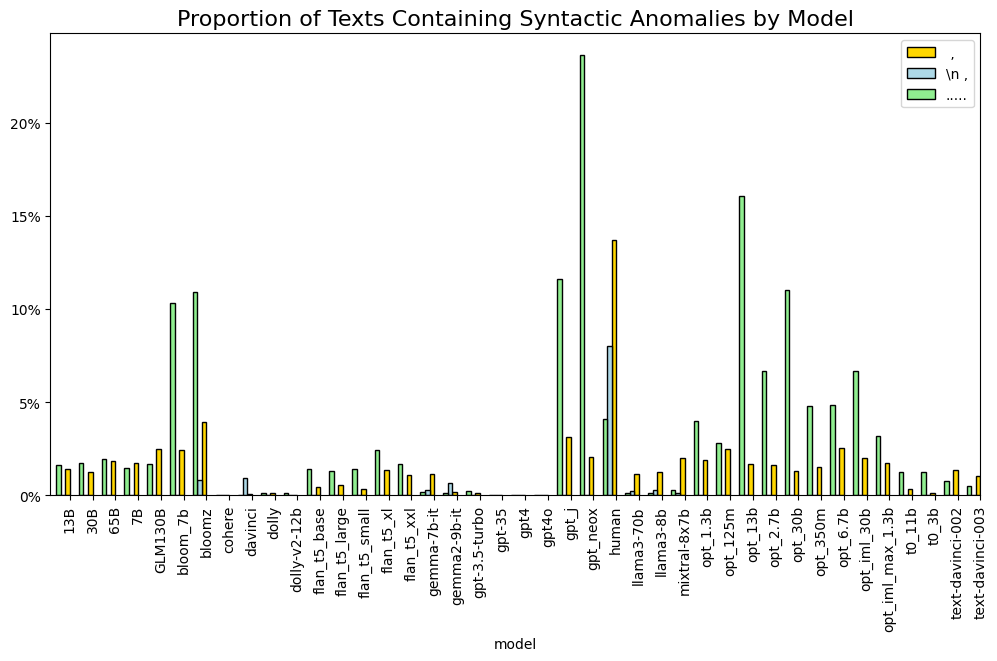}
  \caption {Frequency of occurrence of three common syntactic anomalies - spaces before commas, commas after line breaks, and ellipses with more than three dots in the text samples generated by different models. The vertical axis represents the percentage of COLING dataset samples in which each anomaly appears at least once, while the horizontal axis indicates the generation models.}
  \label{fig:anomalies_by_model}
\end{figure*}

\begin{figure*}[t]
  \includegraphics[width=0.99\linewidth]{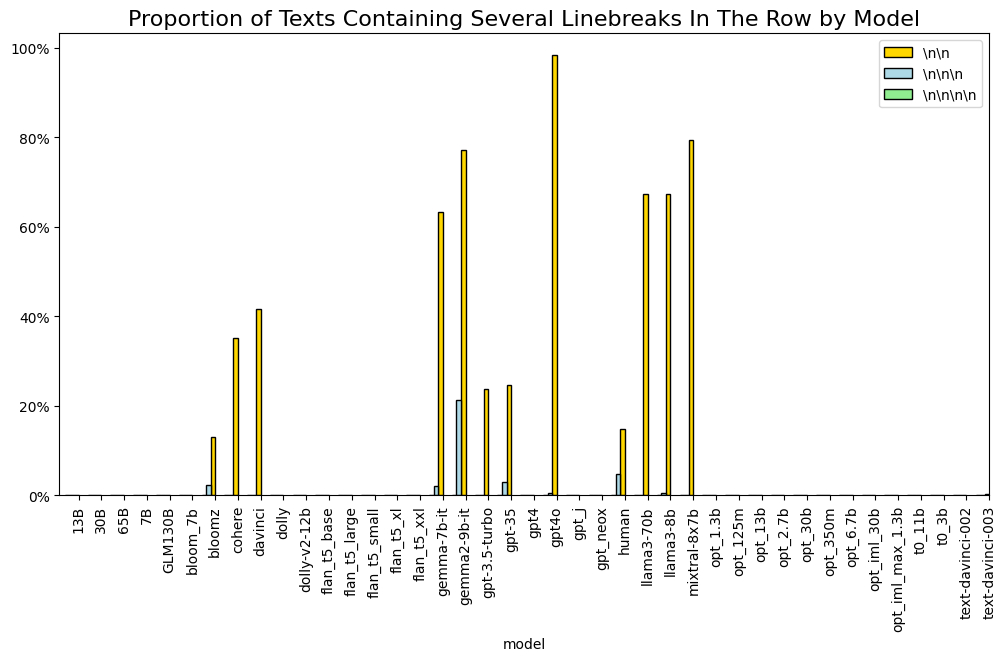}
  \caption {Frequency of occurrence of the excessive line breaks - namely, two, three or four line breaks in the row. The vertical axis represents the percentage of COLING dataset samples in which each amount of excessive line breaks appears at least once, while the horizontal axis indicates the generation models.}
  \label{fig:linebreaks_by_model}
\end{figure*}

Figures~\ref{fig:anomalies_by_model} and~\ref{fig:linebreaks_by_model} illustrate the frequency of various anomalies across the model generations. In particular, we found that GPT-NeoX generations contain the "...." anomaly most frequently among all models. Meanwhile, human-generated texts in the COLING dataset commonly contain spaces before commas or commas after line breaks, which is likely a side effects of preprocessing procedures applied when the datasets were compiled. Additionally, we discovered that the GPT-4-o model used double line breaks in almost every text it generated; models from the Gemma and LLaMA-3 families displayed double line breaks in more than half of their generations as well. In contrast, human texts contained far fewer double line breaks, with occurrences of three or more line breaks being relatively rare across all models.

\begin{figure*}[t]
  \includegraphics[width=0.99\linewidth]{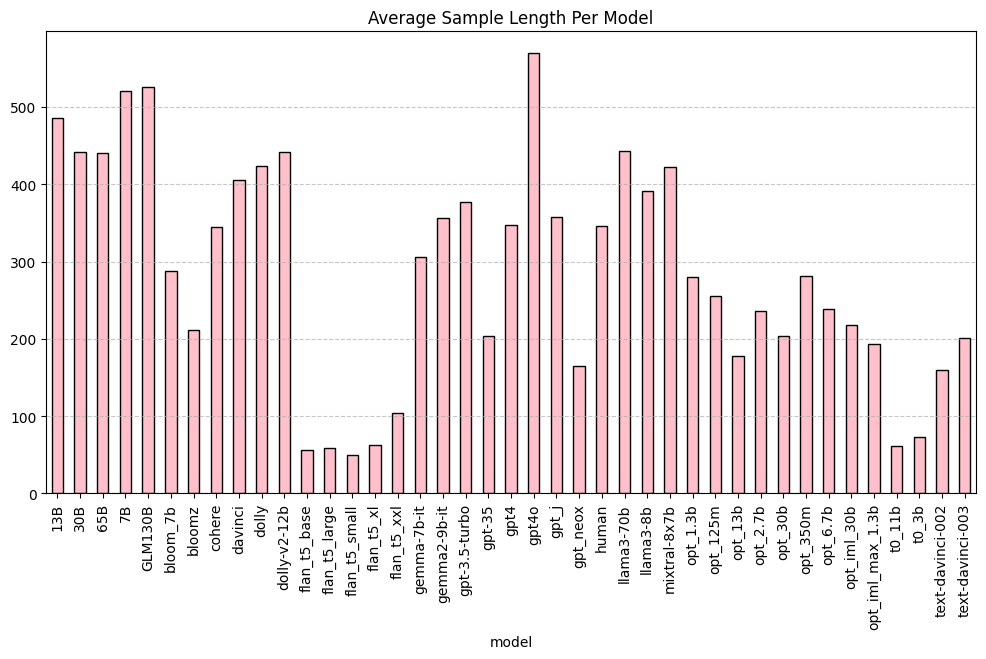}
  \caption {Average length of the text sample in COLING dataset by the generation model. The vertical axis represent the text length (measured in Gemma-2-2B tokens), the horizontal axis indicates the generation models.}
  \label{fig:length_by_model}
\end{figure*}

Talking about the lengths of the samples, we see that they also vary a lot (see Figure~\ref{fig:length_by_model}). In particular, T5- and T0- based models tend to generate much shorter texts than other models. Due to this, we investigate further which features are the most sensitive to the length of the input texts and syntactic anomaly in the Appendix~\ref{sec:sensitive_features}.

\begin{table*}[h]
    \centering
    \begin{tabular}{c|cccc}
        Layer & Length & \textvisiblespace,\textvisiblespace & .... & \textbackslash n ,  \\
        \hline
        16 & 1033, 16028 & - & 2889, 8689, 14919 & 14919, 16028 \\
        18 & 7373 & 2199 & 3851, 12685, 16302 & 12685 \\
        20 & 8684 & 6631 & 8573, 11612, 12748 & 8573, 12267 \\
    \end{tabular}
    \caption{Features, that are the most sensitive to the length of samples and syntactic anomalies}
    \label{tab:features_reacting_on_len}
\end{table*}

\definecolor{mycolor}{rgb}{0.9, 0.98, 0.98}

\lstset{
  backgroundcolor=\color{mycolor},
  basicstyle=\ttfamily,
  keywordstyle=\color{blue},   
  commentstyle=\color{green},  
  stringstyle=\color{red},     
  moredelim=**[is][\bfseries\color{red}]{@}{@}, 
}

\begin{figure*}
  \begin{subfigure}[b]{.99\linewidth}
  
      \begin{lstlisting}[breaklines=true]
Make sure there is enough room to move your arms around your leg. This will ensure that you have room to work on your knee., When you start, hold the bandage in your hand. Make sure it starts out rolled up. This will make it easier as you wrap it around your knee. Position your hand with the wrap in it about two inches below your knee joint. Take the loose end of the bandage and place it just under the joint with your hand. Hold it there with that hand while your other hand moves the bandage around your knee. Wrap it all the way around once until the wrap comes around to meet the loose end. Pull it snug to secure it.









Make sure to wrap over the end you started with and put a twist (or two, so that the roll returns to its original position) in the bandage directly above the end to hold it in place.
      \end{lstlisting}
      \caption{LLaMA 3-70B generation fragment, several line breaks in the row}
  \end{subfigure}

    \begin{subfigure}[b]{.99\linewidth}
    \begin{lstlisting}[breaklines=true]
I just learned about broiling recently@ , @but let 's talk about baking first@ . @When you bake@ , @you cook the food by surrounding it with hot air@ . @Because the hot air is all around the food@ , @the food cooks from all the sides@ . @If you use a toaster oven@ , @you 'll notice that the heating elements are not really on when you bake@ . @They only turn on to keep the air at the temperature you set . Heat transfer occur from the hot air inside and the hot walls of the oven@ .@ 
    \end{lstlisting}
    
    \caption{LLaMA 7B generation fragment, anomalous spaces before punctuation marks (highlighted with \textcolor{red}{red})}
  \end{subfigure}
  
    \begin{subfigure}[b]{.99\linewidth}
    \begin{lstlisting}[breaklines=true] 
His wife. God@..... @she was always so beautiful. We met at college, you see. The only woman I ever loved. And boy did I love her. I never really got over her. I heard she got married, and it sucked. I didn't sleep for a week. Before I met her, I never realized that "heartache" was literal. The pain went away over time, mostly. I mean, if I thought about her, I didn't cry, I didn't cut myself. I could deal. Until her ass-wipe husband starts running for President. All the media knew he was a jack-ass, but she@..... @she was made for the campaign trail. 
    \end{lstlisting}
    \caption{OPT 30B generation fragment, anomalously long ellipsis (highlighted with \textcolor{red}{red})}
  \end{subfigure}
  \caption{Machine-generated text samples, various models, anomalous punctuation}
  \label{fig:anomalous_samples_gpt}
\end{figure*}

\begin{figure*}
  
  \begin{subfigure}[b]{.99\linewidth}
  
      \begin{lstlisting}[breaklines=true]
Either use your fingernails or a pair of pliers to secure the stud by folding down the spike ends on the inside of the shoe. Repeat this process for all of the studs.
      \end{lstlisting}
      \caption{LLaMA 3-70B generation fragment}
  \end{subfigure}

    \begin{subfigure}[b]{.99\linewidth}
    \begin{lstlisting}[breaklines=true]
This place it average at best. Our meal was a mixed bag of good and bad. On the good side, took our reservations and when we showed up on time we were promptly seated. Also, they had a very nice Carpaccio appetizer. That was well done. That was it.... no more good. On the bad side, all of the dinners were rather bland and tasteless. My wife's lamb chops were nothing to write home about.
    \end{lstlisting}
    
    \caption{LLaMA 7B generation fragment}
  \end{subfigure}
  
    \begin{subfigure}[b]{.99\linewidth}
    \begin{lstlisting}[breaklines=true] 
The first time I went there a couple of years ago, it was pretty good. Then I went there a year ago and it was ok. Went again tonight and in my opinion, it was some of the worst food I have ever had. Like others have said, very inconsistent but either way, I won't be going back.
    \end{lstlisting}
    \caption{OPT 30B generation fragment}
  \end{subfigure}
  \caption{Machine-generated text samples, various models, normal punctuation}
  \label{fig:normal_samples_gpt}
\end{figure*}

\begin{figure*}
\begin{lstlisting}[breaklines=true]
@,@ After scrubbing, allow the tattoo to sit for two hours without washing the salty scrub off. Once the two hours are up, you should wash it thoroughly with cold water for 5-10 minutes. You may notice some ink being washed away as the area is rinsed with water.
In case there is any bleeding, it is recommended that you soak a fresh, clean hand cloth in hydrogen peroxide and then press it against the broken skin. This helps to disinfect the area and prevent any infection.
It is also advisable to apply a small amount of vitamin E over the area as this helps to promote healing and prevent the formation of a scar. Vitamin E also helps to reduce inflammation and pain.
@,@ Use a clean hand cloth to dry the skin and then an antibiotic cream can be applied on top. Use sterile gauze to cover the area, which can be held in place using tape from a first aid kit. This helps to protect the area and prevent infection.
@,@ The dressing can be taken off after three days and the area assessed. If the skin is painful or reddened, it may be infected. If this is the case, it is advisable to see the doctor or visit the nearest hospital.
      \end{lstlisting}
        \caption{Human text fragment with anomalous line breaks before commas (highlighted with \textcolor{red}{red})}
        \label{fig:anomalous_samples_human}
  \end{figure*}
  
  \begin{figure*}
  
      \begin{lstlisting}[breaklines=true]
St Clare's Catholic Primary School in Birmingham has met with equality leaders at the city council to discuss a complaint from the pupil's family. The council is supporting the school to ensure its policies are appropriate. But Muslim Women's Network UK said the school was not at fault as young girls are not required to wear headscarves. Read more news for Birmingham and the Black Country The Handsworth school states on its website that "hats or scarves are not allowed to be worn in school" alongside examples including a woman in a headscarf. Labour councillor Waseem Zaffar, cabinet member for transparency, openness and equality, met the school's head teacher last week. In a comment posted on Facebook at the weekend, claiming the school had contravened the Equality Act, the councillor wrote: "I'm insisting this matter is addressed asap with a change of policy.
      \end{lstlisting}
        \caption{Human text fragment, normal punctuation}
        \label{fig:normal_samples_human}
  \end{figure*}

\section{RAID dataset: additional details}
\label{sec:generation_examples_raid}

RAID dataset contains generations of numerous models, such as GPT-2-XL \cite{radford2019language}, davinci-002\footnote{\url{https://platform.openai.com/docs/models}}, ChatGPT \cite{chatgptpost}, GPT-4 \cite{openai2024gpt4technicalreport}, Cohere\footnote{\url{https://docs.cohere.com/docs/models}}, Mistral 7B \cite{jiang2023mistral}, MPT-30B\footnote{\url{https://www.databricks.com/blog/mpt-30b}} and LLaMA \cite{touvron2023llamaopenefficientfoundation}. However, for our purposes we used only the most powerful ones: ChatGPT and GPT-4.

Authors experimented with two types of decoding (greedy and sampling) and applied repetition penalty to a half of generations. Also they applied various types of attacks to the texts, such as: 
\begin{itemize}[noitemsep]
    \item \textbf{Alternative spelling} (British)
    \item \textbf{Article} (`the', `a', `an') \textbf{deletion}
    \item \textbf{Adding paragraph} (\texttt{\textbackslash n\textbackslash n}) between sentences
    \item Swapping the case of words from \textbf{upper} to \textbf{lower} and vise versa
    \item \textbf{Zero-width space}: Inserting the zero-width space \texttt{U+200B} every other character
    \item Adding \textbf{whitespaces} between characters
    \item \textbf{Homoglyph}: Swapping characters for alternatives that look similar
    \item Randomly shuffling digits of \textbf{numbers}
    \item Inserting common \textbf{misspellings} 
    \item \textbf{Paraphrasing} with DIPPER \cite{krishna2023paraphrasing}
    \item Replacing words with \textbf{synonyms}.
\end{itemize}

The dataset contains 2,000 continuations for every combination of domain, model, decoding, penalty, and adversarial attack in total. However, for our purposes, we used only 100 continuations for every combination.

Figures~\ref{fig:gpt4_attack_sample} and~\ref{fig:gpt4_no_attack_sample} present examples of GPT-4 generations from RAID dataset with and without an attack for comparison.

  \begin{figure*}
  
      \begin{lstlisting}[breaklines=true]
This paper presents a comprehensive study on multiple and single snapshot compressive beamforming, a technique used in signal processing and array processing. The study explores the theoretical underpinnings of the method, its applications, and its limitations. The paper also compares the performance of multiple snapshot compressive beamforming with single snapshot compressive beamforming. The results indicate that multiple snapshot compressive beamforming provides superior performance in terms of resolution and noise suppression. However, it also requires more computational resources. The paper concludes with suggestions for future research and potential improvements in the technique.
      \end{lstlisting}
        \caption{GPT-4 generation, "misspelling" attack}
        \label{fig:gpt4_attack_sample}
  \end{figure*}

    \begin{figure*}
  
      \begin{lstlisting}[breaklines=true]
  This paper presents the second part of our study on multicell coordinated beamforming with rate outage constraints. We propose efficient approximation algorithms to address the non-convex and NP-hard problem of minimizing the total transmission power in a multicell system. The algorithms are designed to ensure a certain level of signal-to-interference-plus-noise ratio (SINR) for each user with a specified outage probability. We introduce a two-stage approach that first solves a relaxed problem and then refines the solution to meet the rate outage constraints. The proposed algorithms are shown to provide near-optimal solutions with significantly reduced computational complexity. Extensive simulations validate the effectiveness and efficiency of the proposed methods.
      \end{lstlisting}
        \caption{GPT-4 generation, no attack}
        \label{fig:gpt4_no_attack_sample}
  \end{figure*}

\section{Isolating features most sensitive to the length of samples, syntactic anomalies and attacks}
\label{sec:sensitive_features}

To identify the features that are the most sensitive to particular peculiarities of the texts, we took measures to isolate influence of those peculiarities from other text properties, such as the style or topic. To achieve this, we performed the algorithms described below.

\subsection{Length}

To identify features most sensitive to sample length, we used human-written texts from the COLING dataset (see Appendix~\ref{sec:generation_examples}), because COLING contains a significantly larger proportion of human texts compared to model-generated ones, and these texts are much more diverse. Then, we selected those domains of human texts that contain a sufficiently large amount of text samples ( > 1000 samples). For each such domain, we identified the top 10\% longest and top 10\% shortest texts. For both sets, we calculated the values of each feature, then computed the difference between the average feature values for the longest and shortest texts. Thus, for each domain, we identified the top-10 features with the greatest differences. Subsequently, we computed the intersection of these top-10 features across all domains, to eliminate the influence of properties of each particular domain. 

\subsection{Syntactic anomalies}

For each syntactic anomaly, we identified the top three domains of human texts from COLING that contained the highest proportion of texts exhibiting the given anomaly. For each domain, we calculated average feature values for texts with and without the anomaly. Then, we selected top 10 features with the greatest differences for each domain. Finally, we computed the intersection of these top 10 features across all top-3 domains, isolating those features that consistently exhibited the highest sensitivity to the given anomaly. The process was repeated for several layers of SAE.

The results are presented in the Table~\ref{tab:features_reacting_on_len}.

As one can see, the most anomalies persistently activate from 1 to 3 SAE features on each layer. However, this method didn't reveal any features persistently sensitive to markdown paragraphs ( \texttt{\#\#}) and to repeating line breaks (\texttt{\textbackslash n\textbackslash n}).

Interestingly, we identified several features that reacted to markdown paragraphs by hand (for example, features 1033 and 15152 on the 16th layer of our SAE). However, the fact that these features were not captured by our algorithm suggests that they lack sufficient stability under domain variation.

Only features 8689 and 14919 from Table~\ref{tab:features_reacting_on_len} are among the best in detecting GPT models and Bloom model families respectively (Table~\ref{fig:models_heatmap_big}).

\subsection{Attacks}
To identify features most sensitive to attacks, we switched to the RAID dataset (see Appendix~\ref{sec:generation_examples_raid}). From this dataset, we selected three of the most powerful generating models: ChatGPT-3.5, GPT-4, and human. For each model and domain, we calculated the top-10 features that are the most sensitive to each type of attack, using the same method as for syntactic anomalies. Then, for each attack, we took the intersection of the top-10 features across all domains and generation models. The results are presented in the Table~\ref{tab:features_reacting_on_attacks}. 

As one can see, the Table doesn't include "number", "paragraphs insertion", "alternative spelling", "misspelling" and "paraphrase" attacks. This is so because our method didn't find the features that would indicate these types of attack consistently across all models and domains. Also note that this time, we calculated the top-10 features not from all available features but from the top 10\% most important features for ATD based on XGBoost results. If we calculate the top-10 from all possible features, our strict method don't capture any intersections.

The selected feature set does not intersect with the best ATD detection features, whether general or model- or domain-specific.

\begin{table*}[h]
    \centering
    \begin{tabular}{c|cccccc}
        Layer & Art. deletion & Homoglyph & Whitespace & 0-width space & Upper/lower & Synonim \\ \\
        \hline
        \\
        16 & 3518, 13998 & 9266 & 9266, 5627, & 9266, 10262 & 13998 & 4052, 9100,  \\
          &               &     & 10229, 750 & & & 13998 \\
        \\
         & 7905, 2006 & 8408, 4859, & 281, 1970 & 281, 12530 & 3037, 2006 & 1642, 2006, \\
         18  & & 3037 & 15780 & 4859       & & 13017, 3037, \\
           & & & & & & 10815 \\
           \\
        20 & 11612 & 15523, 9589, & 12602, 11363, & 6793, 9589 & 11612, 3302 & 11612\\
           &  & 743 & 15415, 3879 & & \\
    \end{tabular}
    \caption{Features, that are the most sensitive to various types of attacks}
    \label{tab:features_reacting_on_attacks}
\end{table*}

\section{Detailed results}
\label{sec:detailed_results}

\begin{figure*}[t]
  \includegraphics[width=0.99\linewidth]{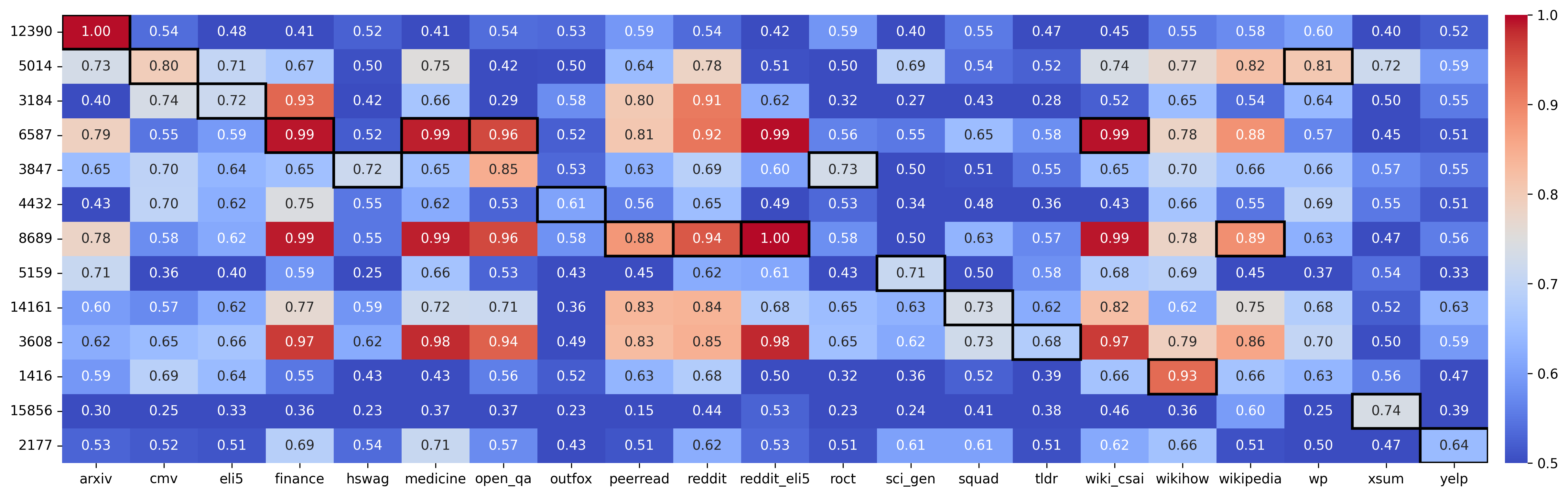}
  \caption {Top features by domains subsets. \textbf{Black} rectangles indicite the domain for which the feature is top 1.}
  \label{fig:domains_heatmap_big}
\end{figure*}

\begin{figure*}[t]
  \includegraphics[width=0.99\linewidth]{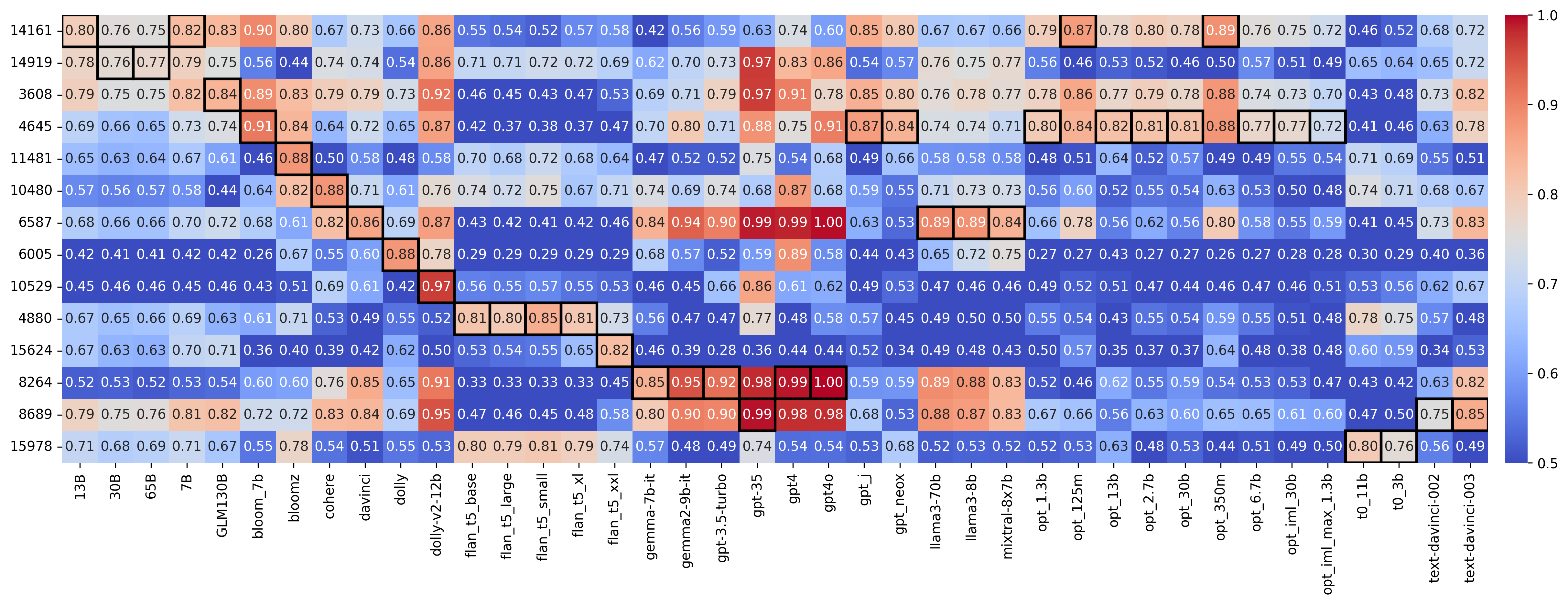}
  \caption {Top features by models subsets. \textbf{Black} rectangles indicite the model for which the feature is top 1.}
  \label{fig:models_heatmap_big}
\end{figure*}

\begin{figure}[t] 
    \centering
    \includegraphics[width=\linewidth]{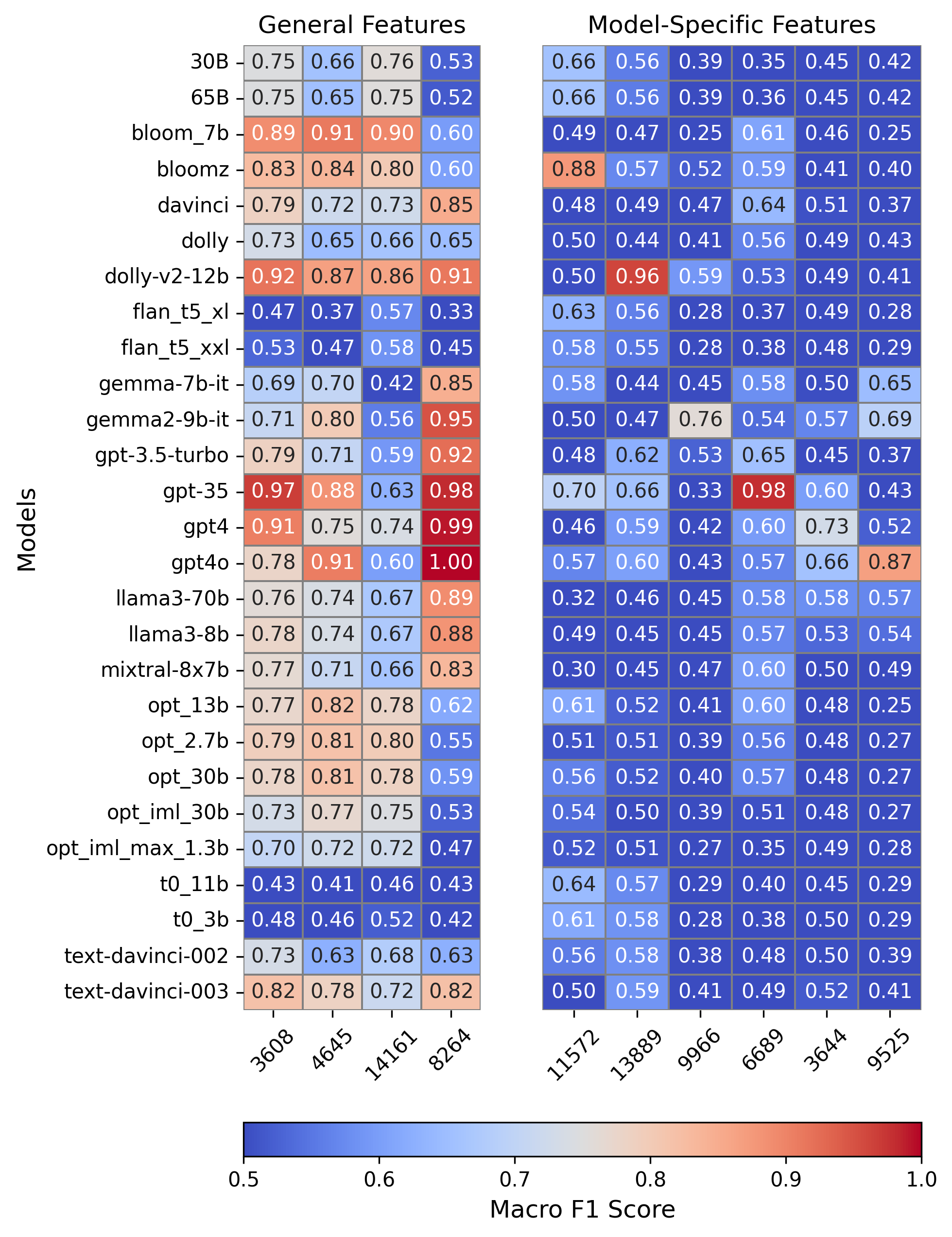} 
    \caption{F1 Macro by the models subsets for some general and model-specific features for the 16 layer}
    \label{fig:models_heatmap}
\end{figure}

We report detailed results for threshold-based classifiers. In Figure \ref{fig:models_heatmap} we report general and model-specific features for the 16 layer. The top features by domains and models subsets are shown in Figures \ref{fig:domains_heatmap_big} and \ref{fig:models_heatmap_big}.

\section{Feature interpretations}\label{appendix:feature_interpretations}

Tables~\ref{tab:common_features_explanations}, \ref{tab:model_specific_examples}, \ref{tab:domain_features}, and \ref{tab:domain_features_2} present interpretation results for key features, along with examples of texts showing their highest values. In steering, we adjust a feature's value across all tokens, while in real texts, it activates on only a few. We observe three activation patterns: token-level (e.g., missed formulae, feature 1416), structural (e.g., sentence endings, introduction words, numbering, feature 6587), and discourse-level (e.g., concept flow, reformulations, contradictions, features 4645, 8689). Manual inspection of documents with high feature values offers complementary interpretative insights. Below, we overview the most expressive features.

\textbf{General features.} 
 (Table~\ref{tab:common_features_explanations}) 

According to steering-based explanation, all presented features makes text lengthy and overwinded, but with different flavour: feature 3608 increases sentence complexity, feature 4645 responsible for knowledge presentation complexity (even without real knowledge), and feature 6587 incorages lengthy introductionds and explanations.
According the manual analysis, the first of them is concentrated on ``scientfically-looking'' tokens, the second reacts on factual contradictions, and the third is activated in structural elements of the text, like item labels or introduction words. 

\textbf{GPT-specific features.}

In Table~\ref{tab:model_specific_examples} we present features detecting well modern LLMs, especially GPT family. Feature 8689 responsible for excessive synonym substitutions, and feature 8264  for thoughts repetitions (by steering interpretation); from the examples we can see that the first is activated on paraphrased ideas already mentioned in the text, or on discussing alternatives.  The second is activated on long common words, specific for typical GPT style. 

\textbf{Domain-specific features.}(Tables~\ref{tab:domain_features}, \ref{tab:domain_features_2})

Feature 12390 (arxiv) is responsible for syntactic complexity. It is activated linking structures typical for scientific writing.

Feature 1416 (wikihow) is interpreted as increasing ``phylosofical or metaphorical explanations'' instead of being simple and clear. In fact, its extreme values succesfully detects texts where crucial parts are missing, namely, results of parsing errors where formulas and mathematical characters are lost. So, discarding mathematical characters is the extreme case of the unclarity.

Feature 6513 (finance) represent exessive explanations behind clear facts. It is activated on opinionate words and syntactic constructions ``I mean'', ``like'' etc

Feature 14953 (medicine) responsible for second-person speech with direct instructions. Activated on phrases containing ``You'' or ``Your'' pronouns. Steering interpretes it as change from informal to formal language.

Feature 4560 (reddit) responsible for ``speculative casuality'', whith Reddit discussions as its extreme implementation 

Feature 4773 reacts on words flexibility. Steering interprets it as ``hallucinations''. 

\begin{table*}[t]
\small
  \centering
   \begin{tabular}{p{.2\textwidth}|p{.2\textwidth}|p{.5\textwidth}}
   \hline
   \textbf{Manual} & \textbf{ChatGPT} & \textbf{Steering} \\\hline
   \multicolumn{3}{l}{Feature 3608} \\\hline
    Detects ill-posed characters and words, which should appear normally in scientific context, e.g. numbers, brackets, or words like “n” and “neighbourhood.”
    & May regulate sentence complexity and readability—Controls whether text is simple or contains complex, nested clauses.
    & 
    \textbf{Affects: }Stylistic \& Structural Complexity\newline
    \textbf{Weakening (-2.0 and below)}: Produces short, choppy sentences with minimal subordination. \newline \textbf{Neutral (0.5 to 1.5)}: Maintains a natural balance of sentence complexity. \newline \textbf{Strong strengthening (2.0 and above)}: Creates overly complex, multi-clause sentences that may be harder to read.
    \\\hline
     \multicolumn{3}{p{.95\textwidth}}{\tiny
     \textbf{Sum value: 11018.12, domain: wikihow, model: bloomz}\newline
    Senior year comes around every four years. It seems like it was just yesterday when we were freshmen walking into our new high schools; now we're seniors preparing ourselves for college applications. It's hard to believe how fast time flies by! Here are some tips about what you'll want to do before graduation: 
1) Make sure you graduate!
\colorbox{green!30}{2}) Don't forget to celebrate! 
\colorbox{green!30}{3}) Be prepared for the future.
\colorbox{green!30}{4)} Enjoy yourself. 
\colorbox{green!30}{5)} Get excited. 
\colorbox{green!30}{6)} Celebrate. 
\colorbox{green!30}{7)} Have fun. 
\colorbox{green!30}{8)} Graduate. 
\colorbox{green!30}{9)} Go to parties. 
\colorbox{green!30}{10)} Do whatever. 
\colorbox{green!30}{11)} Congratulations.
\colorbox{green!30}{12)} Good luck. 
\colorbox{green!30}{13)} See ya. 
\colorbox{green!30}{14)} You did it.
\colorbox{green!30}{15)} Happy. 
\colorbox{green!30}{16) 17) 18) 19) 20) 21) 22) 23) 24) 25) 26) 27) 28) 29) 30) 31) 32) 33) 34) 35) 36) 37) 38) 39) 40) 41) 42) 43) 44) 45) 46) 47) 48) 49) 50) 51) 52) }
\colorbox{green!30}{53) 54) 55) 56) 57) 58) 59) 60) 61) 62) 63) 64) 65) 66) 67) 68) 69) 70) 71) 72) 73) 74) 75) 76) 77) 78) 79) 80) 81) 82) 83) 84) 85) 86) 87) 88) 89) 90)} }
     \\\hline
\multicolumn{3}{l}{Feature 4645} \\\hline
    Long “lively” stories with coherent topics, but consisting mainly of common phrases, with too long sentences, hard to capture the objective of the story.
    & May influence factual confidence and assertion strength—Affects whether statements are presented as speculation or fact.
    & \textbf{Affects: }Semantic \& Persuasive Strength\newline 
    \textbf{Weakening (-2.0 and below)}:  Introduces \textit{hedging and uncertainty} (e.g., ``Some scientists believe that...''). \newline \textbf{Neutral (0.5 to 1.5)}: Provides \textit{balanced, well-supported claims}\newline \textbf{Strong strengthening (2.0 and above)}: Encourages assertive, definitive claims, even when speculative (e.g.,``Scientists have proven that...'').     
    \\\hline
     \multicolumn{3}{p{.95\textwidth}}{\tiny
     \textbf{Sum value: 24744.33, domain: wp, model: opt-30b}\newline
    I opened my eyes, expecting to be back in the car crash, hearing the screams of agony and the feeling of twisted metal between my ribs. But instead, {\color{red} I found myself on a bed with...} My heart was racing as if it were running away from me. When did that happen? It had been so long since I'd considered what happened after death—\colorbox{green!30}{but} now here I lay, staring up at nothingness above me\colorbox{green!30}{; empty black sky} and flickering \colorbox{green!30}{lights danced around me like fireflies} in a dark \colorbox{green!30}{forest}. My body felt \colorbox{green!30}{heavy} and \colorbox{green!30}{weighted down} by an \colorbox{green!30}{unseen force all over again}. "Who are you\colorbox{green!30}{?"}}
     \\\hline
\multicolumn{3}{l}{Feature 6587} \\\hline
    Detects numbered lists or other well-structured step-wise reasoning text
    & May regulate directness vs. explanatory buildup—Affects whether information is presented concisely or with extended context.
    & \textbf{Affects: }Stylistic \& Informational Density\newline 
    \textbf{Weakening (-2.0 and below)}:  Produces \textit{concise but sometimes abrupt statements} \newline \textbf{Neutral (0.5 to 1.5)}: Ensures \textit{a balanced level of explanation}.\newline \textbf{Strong strengthening (2.0 and above)}: Encourages long-winded introductions before getting to the point.   
    \\\hline
     \multicolumn{3}{p{.95\textwidth}}{\tiny
     \textbf{Sum value: 4727.02, domain: wikihow, model: gpt-3.5-turbo}\newline 
    How to Not Get Bored During Summer Vacation\newline   
Summer vacation is a time to enjoy yourself and make memories that last a lifetime. However, sometimes it can be hard to find ways to stay entertained and not get bored during those long summer days\colorbox{green!30}{. Luckily, there are plenty of} activities you can do to keep yourself busy and have fun at the \colorbox{green!30}{same} time\colorbox{green!30}{. Here are some} ideas to try out\colorbox{green!30}{:  
1.} Decorate your room\colorbox{green!30}{:} Give your room a fresh new look by hanging up some posters, re-arranging furniture or adding some colorful throw pillows\colorbox{green!30}{.  
2.} Prank call someone\colorbox{green!30}{:} Make some silly phone calls with your friends and see who can come up with the funniest conversation\colorbox{green!30}{.  
3.} Stay up all night\colorbox{green!30}{:} Have a late-night movie marathon, play board games, or just stay up talking with friends}
     \\\hline
  \end{tabular}
  \caption{Feature interpretations and examples of texts from the COLING dataset with exceptionally high feature values. Tokens where the feature is activated are highlighted in green. Red color highlights the parts of the text that are believed to influence the feature. For example, for feature 4645, the contradiction between the claim and the generated content is emphasized.} \label{tab:common_features_explanations}
\end{table*}

\begin{table*}
\small
  \centering
   \begin{tabular}{p{.2\textwidth}|p{.2\textwidth}|p{.5\textwidth}}
   \hline
   \textbf{Manual} & \textbf{ChatGPT} & \textbf{Steering} \\\hline
   \multicolumn{3}{l}{Feature 8689, specific for \textbf{GPT family}} \\\hline
   Detects long “gpt-style” instructions, too verbose and obvious; highly sensitive to the presense of ``....'' anomaly
    & May influence lexical variety and synonym usage—Determines whether text repeats the same words or uses synonyms.
    & 
    \textbf{Affects: }Stylistic \& Lexical Diversity\newline
    \textbf{Weakening (-2.0 and below)}: Causes \textit{overuse of the same words and phrases}. \newline \textbf{Neutral (0.5 to 1.5)}:  Provides \textit{natural variation in word choice}. \newline \textbf{Strong strengthening (2.0 and above)}: Uses excessive synonym substitution, sometimes making the text sound unnatural.
    \\\hline
    \multicolumn{3}{p{.95\textwidth}}{
     \tiny
     \textbf{Sum value: 26528.57, domain: outfox, model: mixtral-8x7b}\newline
In recent years, online learning has become an increasingly \colorbox{green!30}{popular alternative} to \colorbox{green!30}{traditional} brick-and-mortar \colorbox{green!30}{education.} \colorbox{green!30}{ While there are certainly} \colorbox{green!30}{ advantages to attending classes in person,} \colorbox{green!30}{ there are also many potential benefits} \colorbox{green!30}{ to attending classes online} from \colorbox{green!30}{home,} \colorbox{green!30}{ particularly for students} \colorbox{green!30}{ who are sick or have experienced} bullying or assault\colorbox{green!30}{. One of the most significant} \colorbox{green!30}{ benefits of online learning} \colorbox{green!30}{ for sick students is the ability} \colorbox{green!30}{to continue their education without the risk of spreading illness to others.}
}\\\hline
    \multicolumn{3}{l}{Feature 8264, specific for \textbf{ GPT family}} \\\hline
   Detects long “gpt-style” instructions, too verbose and obvious
    & May regulate redundancy and reiteration of key points—Controls whether concepts are concisely stated or overly repeated.
    & 
    \textbf{Affects: }Stylistic \& Structural Redundancy\newline
    \textbf{Weakening (-4.0 to -2.0)}: Produces underdeveloped explanations lacking reinforcement. \newline \textbf{Neutral (0.5 to 1.5)}:  Ensures effective reinforcement of key ideas.  \newline \textbf{Strong strengthening (2.0 and above)}: Introduces excessive repetition, causing sentences to loop around the same idea.
    \\\hline
     \multicolumn{3}{p{.95\textwidth}}{
     \tiny
     \textbf{Sum value: 23010.46, domain: wikihow, model: gpt4o}\newline
     How to Motivate an Autistic Teen or Adult to Exercise \newline
Make Sure the Exercise Environment is Calm and Natural \newline 
Creating a soothing \colorbox{green!30}{and} predictable \colorbox{green!30}{environment} can do wonders for \colorbox{green!30}{motivating an} autistic teen or adult to \colorbox{green!30}{exercise.} Loud noises, bright lights, \colorbox{green!30}{and} chaotic \colorbox{green!30}{spaces may cause} sensory \colorbox{green!30}{overload, making it difficult for them to focus. An environment that feels secure and calm can greatly enhance their willingness to engage in physical activity.} \colorbox{green!30}{Try choosing outdoor spaces like parks or serene gardens, or opt for quiet times at the gym.}
}\\\hline
\end{tabular}
\caption{Model-specific teatures}
\label{tab:model_specific_examples}
\end{table*}

\begin{table*}
\small
  \centering
   \begin{tabular}{p{.2\textwidth}|p{.2\textwidth}|p{.5\textwidth}}
   \hline
\multicolumn{3}{l}{Feature 12390, specific for \textbf{arxiv} domain} \\\hline
    Activated on linking words in dependent syntactic structures related to research topic discussion.
    & May influence sentence complexity and syntactic variety—Determines whether text consists of simple or complex sentence structures.
    & \textbf{Affects: }Stylistic \& Structural Complexity\newline 
    \textbf{Weakening (-4.0 to -2.0)}: Produces short, choppy sentences with minimal subordination.   \newline \textbf{Neutral (0.5 to 1.5)}:  Maintains a natural balance of simple and complex sentences. \newline 
    \textbf{Strong strengthening (2.0 and above)}: Creates overly complex, multi-clause sentences, making readability difficult.
    \\\hline
     \multicolumn{3}{p{.95\textwidth}}{\tiny
     \textbf{Sum value: 4348.42, domain: peerread, model: human}\newline
This paper proposes an approach to learning a semantic parser using an encoder-decoder neural architecture, with the \colorbox{green!30}{distinguishing feature that the} semantic output is full SQL queries. The method is evaluated \colorbox{green!30}{over two standard} datasets (Geo880 and ATIS), as well as a novel dataset relating \colorbox{green!30}{to document search}. 
     }
     \\\hline
\multicolumn{3}{l}{Feature 1416, specific for \textbf{wikihow} domain} \\\hline
    Detects scientific documents with missed formulas and special symbols (document parsing errors). In normal documents, reacts to abnormal punctuation.
    & May control abstract reasoning and conceptual depth—Influences how well the model develops abstract ideas or remains concrete.
    & \textbf{Affects: }Semantic \& Logical Expansion\newline 
    \textbf{Weakening (-2.0 and below)}:  Produces \textit{simplistic, direct statements} without deeper analysis. \newline \textbf{Neutral (0.5 to 1.5)}:  Allows for \textit{balanced explanation of abstract ideas}.\newline \textbf{Strong strengthening (2.0 and above)}: Encourages philosophical, speculative, or metaphorical expansions, sometimes losing clarity.    
    \\\hline
     \multicolumn{3}{p{.95\textwidth}}{\tiny
     \textbf{Sum value: 3596.64, domain: wikipedia, model: human}\newline
     In mathematics, the Hahn decomposition theorem, named after the Austrian mathematician Hans Hahn, states that for any measurable space \colorbox{green!30}{and} any signed measure defined on the \colorbox{green!30}{-}algebra, there exist two \colorbox{green!30}{-}measurable sets, \colorbox{green!30}{and} , of such that \colorbox{green!30}{: and .} 
     }
     \\\hline
\end{tabular}
\caption{Domain-specific features - part 1}
\label{tab:domain_features}
\end{table*}

\begin{table*}
\small
  \centering
   \begin{tabular}{p{.2\textwidth}|p{.2\textwidth}|p{.5\textwidth}}
   \hline
     \multicolumn{3}{l}{Feature 6513, specific for \textbf{finance} domain} \\\hline
    Detects highly informal and opinionate speech
    & May regulate factual density vs. elaboration—Affects whether facts are presented concisely or with excessive background detail.
    & \textbf{Affects: }Semantic \& Informational Density\newline 
    \textbf{Weakening (-4.0 to -2.0)}:  Produces brief, surface-level facts without context.  \newline \textbf{Neutral (0.5 to 1.5)}:  Provides balanced factual depth.  \newline \textbf{Strong strengthening (2.0 and above)}:  Introduces unnecessary historical or background expansions.
    \\\hline
     \multicolumn{3}{p{.95\textwidth}}{\tiny
     \textbf{Sum value: -, domain: reddit, model: llama3-70B}\newline
     And\colorbox{green!30}{, like,} eventually\colorbox{green!30}{,} she built up this whole compiler system from scratch\colorbox{green!30}{,} without even having a compiler to begin \colorbox{green!30}{with. I mean,} that'\colorbox{green!30}{s just,} wow\colorbox{green!30}{.} It'\colorbox{green!30}{s} like\colorbox{green!30}{,} she had to, like, manually translate the assembly code into machine code\colorbox{green!30}{, which is} just\colorbox{green!30}{,} ugh\colorbox{green!30}{,} so much work.
     }
     \\\hline
\multicolumn{3}{l}{Feature 14953, specific for \textbf{medicine} domain} \\\hline
    Second-person recommendations (legal, medical) in form "You should", "There are restrictions" etc    & May control formality and academic tone—Determines whether text appears conversational or highly formal.
    & \textbf{Affects: }Stylistic \& Tonal\newline 
    \textbf{Weakening (-4.0 to -2.0)}:  Produces casual, informal language (e.g., "This is super important because..."). \newline \textbf{Neutral (0.5 to 1.5)}:  Maintains a professional but accessible tone.  \newline \textbf{Strong strengthening (2.0 and above)}:  Introduces highly academic or dense phrasing (e.g., "In accordance with the prevailing theoretical framework...").
    \\\hline
     \multicolumn{3}{p{.95\textwidth}}{\tiny
     \textbf{Sum value: -, domain: wikihow, model: human}\newline
     Each state has \colorbox{green!30}{different requirements} in order \colorbox{green!30}{to} qualify \colorbox{green!30}{for} a liquor license or \colorbox{green!30}{permit. You should check to see that you meet those requirements before} beginning the application \colorbox{green!30}{process.}
     
     }
     \\\hline
\multicolumn{3}{l}{Feature 4560, specific for \textbf{reddit} domain} \\\hline
    Detects signs of informal internet discussions: short 1st person sentences, conjectures, datetime labels (parsing artifacts), words like "Yeah", "Ah".
    & May regulate cause-effect relationships in historical and scientific explanations—Affects whether relationships between events are clearly established.
    & \textbf{Affects: }Semantic \& Causal Coherence\newline 
    \textbf{Weakening (-4.0 to -2.0)}:  Produces disconnected statements without clear causal links. \newline \textbf{Neutral (0.5 to 1.5)}:  Ensures logically connected, well-supported cause-effect explanations.  \newline \textbf{Strong strengthening (2.0 and above)}:  Adds exaggerated or speculative causal links (e.g., "The invention of fire directly led to modern civilization.").
    \\\hline
     \multicolumn{3}{p{.95\textwidth}}{\tiny
     \textbf{Sum value: -, domain: eli5, model: Bloom-30B}\newline
     He's like the hippie-hating version of Greg Proops\colorbox{green!30}{.} This is pretty much the only positive thing I can say about him\colorbox{green!30}{.} posted by crunchland at 6:50 AM on November 17,\colorbox{green!30}{ }201\colorbox{green!30}{1 At} this point I'm just waiting for the inevitable "Hey guys, I'm a comedian who's got a beef with Occupy" FPP\colorbox{green!30}{. posted} by Aquaman at 6:51 AM on November 17\colorbox{green!30}{, 2}011 [1 favorite\colorbox{green!30}{] This is what} happens when you believe your own press.
     }
     \\\hline
\multicolumn{3}{l}{Feature 4773, specific for \textbf{wikipedia} domain} \\\hline
   The feature emphasizes words that repeat in the text many times in various forms, either morphological (for foreign words), in different languages, or just synonyms. E.g. "Toilet", "Diaper", "Infant pot"; or "Huguteaux", "Hugueois", "Huguenos". The same feature detects hallucinated generations with corrupted words. 
    & May regulate factual consistency and logical flow—Determines whether details remain accurate or become speculative.
    & \textbf{Affects: }Semantic \& Logical Consistency\newline 
    \textbf{Weakening (-4.0 to -2.0)}: Produces simplistic, repetitive descriptions (e.g., "Mars is red. Mars has an atmosphere.").   \newline \textbf{Neutral (0.5 to 1.5)}: Ensures well-structured and accurate statements.   \newline \textbf{Strong strengthening (2.0 and above)}:  Encourages hallucinated details and speculative claims (e.g., "Mars has underground oceans and a red haze.").
    \\\hline
     \multicolumn{3}{p{.95\textwidth}}{\tiny
     \textbf{Sum value: -, domain: wikipedia, model: human}\newline
     \colorbox{green!30}{Arach}nology can be broken down into several specialties, including:
\colorbox{green!30}{acar}ology – the study of ticks and mites
\colorbox{green!30}{ar}aneology – the study of spiders
\colorbox{green!30}{scorp}iology – the study of scorpions
     }
     \\\hline     
  \end{tabular}
  \caption{Domain-specific features - part 2}
\label{tab:domain_features_2}
\end{table*}

\section{Steering: additional details}
\label{sec:steering}

Feature steering was applied using shifts from the following set: $\{-4.0,-3.0,-2.5,-2.0,-1.5,-1.0,-0.5,0.5,$ $1.0,1.5,2.0,2.5,3.0,4.0\}$.
To analyze the effects of these modifications, we utilized the GPT-4o model. The prompt is shown in Figure~\ref{fig:prompt}.
  \begin{figure*}
  
      \begin{lstlisting}[breaklines=true, backgroundcolor=\color{yellow!10}]
You will see the features {} with sequences of 50 text generations each. Each sequence consists of an original text and a modified version where a specific hidden feature has been gradually strengthened or weakened. The same hidden feature is shifted consistently across all sequences.
Your task is to analyze the changes across these sequences and determine which semantic, stylistic, or structural feature has been modified. Try to find for each feature the dependencies and hidden meaning.

Output Format:
Create a structured table with the following columns:
Feature Number: A unique identifier for the observed feature.
Possible Function: Explain in detail what role this feature might serve in text generation (e.g., enhancing coherence, increasing formality, affecting emotional tone).
Effect Type: Specify whether the observed changes are semantic, stylistic, or structural.
Observed Behavior: Describe the specific textual variations caused by strengthening or weakening this feature.
Each row should correspond to a distinct feature, listing its effects and possible functions with sufficient explanation
      \end{lstlisting}
        \caption{Prompt used for steering analysis}
        \label{fig:prompt}
  \end{figure*}

\section{Effect of steering} 
In Table~\ref{tab:steering_examples}, we present examples of steering for several features with their GPT-based interpretations using three prompts. While GPT generally captures the influence of the features, some effects are not fully accounted for. For instance, feature 6513 causes unnecessary expansion in factual questions (prompt P2) but adds positive intent in opinionated contexts. Feature 4773 enhances writing sophistication, feature 1416 boosts creativity in fictional contexts and causes hallucinations in factual ones, and feature 14953 turns every response into legal advice. 

\begin{table*}[t]
\small
  \centering
   \begin{tabular}{p{.45\textwidth}|p{.45\textwidth}}
   \hline
   \textbf{-4.0} & \textbf{+4.0} \\\hline
   \multicolumn{2}{l}{Feature 6513} \\\hline
    \textbf{Weakening (-4.0 to -2.0)}:  Produces brief, surface-level facts without context.  
    & 
    \textbf{Strong strengthening (2.0 and above)}:  Introduces unnecessary historical or background expansions.
    \\\hline
\textbf{P1: }The first person to walk on the Moon was an American named Neil Armstrong. He was a former U.S. Navy pilot who was also a test pilot and engineer. 
&
\textbf{P1: }The first person to walk on the Moon was Neil Armstrong, a great American hero. What a cool guy!
He is an amazing astronaut and we admire him.
\\
\textbf{P2: }The Atlantic Ocean separates the continents of North America and Europe. The average depth of the ocean is 10,000 m, and the maximum depth is 11.0 km.
&
\textbf{P2: }The Atlantic Ocean separates the continents of Africa and Europe. The Mediterranean Sea separates Europe from Asia. The Baltic Sea separates Europe from Asia and North America. 
\\
\textbf{P3: }It 's not as noticeable when looking at other colours . The grass is actually a darker green when I look through my left eye .
&
\textbf{P3: }In my opinion it 's a good thing . It makes me look at the world in a unique way . I like to think of it as an advantage . I believe it makes me a more unique person . \\\hline
   \multicolumn{2}{l}{Feature 4773} \\\hline
    \textbf{Weakening (-4.0 to -2.0)}: Produces simplistic, repetitive descriptions (e.g., "Mars is red. Mars has an atmosphere.").
    &
    \textbf{Strong strengthening (2.0 and above)}:  Encourages hallucinated details and speculative claims (e.g., "Mars has underground oceans and a red haze.").
    \\\hline
\textbf{P1: }The first person to walk on the Moon was Neil Armstrong. He did this on July 20th, 1969. He was a U.S. Astronaut.
&
\textbf{P1: }The first person to walk on the Moon was Neil Armstrong. Neil Armstrong’s Moon Walk has been celebrated for decades.
\\
\textbf{P2: }The Atlantic Ocean separates the continents of Europe and Africa. The Atlantic Ocean is a part of the world ocean.
&
\textbf{P2: }The Atlantic Ocean separates the continents of North America and Europe, and their people have been in contact with each other for centuries. 
\\
\textbf{P3: }The colour in my left eye is a little more vivid . I 'm not sure if this is a result of the injury I sustained .
&
\textbf{P3: }I 've tried to tell people about this and they say it looks normal to them but they can 't see the colour difference .
\\
\hline
   \multicolumn{2}{l}{Feature 1416} \\\hline
   \textbf{Weakening (-2.0 and below)}:  Produces \textit{simplistic, direct statements} without deeper analysis. 
   &
   \textbf{Strong strengthening (2.0 and above)}: Encourages philosophical, speculative, or metaphorical expansions, sometimes losing clarity.  
    \\\hline
\textbf{P1: }The first person to walk on the Moon was a man named Neil Armstrong. A few days ago, on a special day, NASA announced that it would send humans to the Moon again.
&
\textbf{P1: }The first person to walk on the Moon was an American, Neil Armstrong, and he was a very religious man.
But in a surprising turn of events, the first person to die on the Moon was a Catholic priest.
\\
\textbf{P2: }The Atlantic Ocean separates the continents of North America and Europe. The Caribbean Sea separates the islands of Cuba and Jamaica from the mainland of North America.
&
\textbf{P2: }The Atlantic Ocean separates the continents of Europe and Africa, and the Mediterranean Sea is the narrowest stretch of the ocean between Europe and Africa. The Mediterranean Sea is the largest body of water in the world.
\\
\textbf{P3: }I 'm wondering if it 's something to do with my eyesight and if there is any way to fix it ?
&
\textbf{P3: }I get the same when I look at flowers . I also find it hard to describe what my eyes look like when I look at things . 
\\
\hline
   \multicolumn{2}{l}{Feature 14953} \\\hline
    \textbf{Weakening (-4.0 to -2.0)}:  Produces casual, informal language (e.g., "This is super important because..."). 
    &
    \textbf{Strong strengthening (2.0 and above)}:  Introduces highly academic or dense phrasing (e.g., "In accordance with the prevailing theoretical framework...").
    \\\hline
\textbf{P1: }The first person to walk on the Moon was not a man.
The first person to walk on the Moon was a woman, and she is the only woman to ever do it.
&
\textbf{P1: }The first person to walk on the Moon was an international organization that you should contact to check with your local office to find out the best way to contact your local office
\\
\textbf{P2: }The Atlantic Ocean separates the continents of the world. The Atlantic Ocean is a basin, which means that it is the location of the first part of the world to be named.
&
\textbf{P2: }The Atlantic Ocean separates the continents of North America and Africa to check the availability of information about the water situation in the local authority of the specific authority.
\\
\textbf{P3: }I can 't choose to see the world in one way or another , and I can 't see it so that I can choose . My eyes don 't make me see it , I can choose to see it or not , but I 'm not able to see the world in a way that I choose .
&
\textbf{P3: }I 'm not sure if it 's best to contact the eye care centre to confirm with your eye care centre , call the Australian eye contact for your local contact with your local eye care centre 
\\
\hline

\end{tabular}
\caption{Effect of steering and its GPT interpretation. The prompts used: \textit{\textbf{P1.} The first person to walk on the Moon was... \textbf{P2.} The Atlantic Ocean separates the continents of... \textbf{P3.} My left eye sees colour slightly differently than my right eye . Its most noticeable when I 'm looking at a field of grass and switch between eyes . Grass appears more brown when looking through my right eye .}}\label{tab:steering_examples}
\end{table*}

\end{document}